%% file: main.tex
\documentclass[10pt,twocolumn,letterpaper]{article}

\usepackage{wacv}              


\usepackage{graphicx}
\usepackage{subcaption}
\usepackage{enumitem}
\usepackage{tabularx}
\definecolor{wacvblue}{rgb}{0.21,0.49,0.74}
\usepackage[pagebackref,breaklinks,colorlinks,allcolors=wacvblue]{hyperref}
\usepackage{etoolbox}
\patchcmd{\abstract}{\vskip4pc}{\vspace{-0.25em}}{}{}
\usepackage[accsupp]{axessibility}  

\def\wacvPaperID{337} 
\def\confName{WACV}
\def\confYear{2026}

\title{GraspDiffusion: Synthesizing Realistic Whole-body Hand-Object Interaction}

\author{Patrick Kwon\textsuperscript{*}  \\
University of Central Florida\\
{\tt\small yo564250@ucf.edu}
\and
Chen Chen\\
University of Central Florida\\
{\tt\small chen.chen@crcv.ucf.edu}
\and
Hanbyul Joo\\
Seoul National University\\
{\tt\small hbjoo@snu.ac.kr}
}

\begin{document}

\twocolumn[{%
\renewcommand\twocolumn[1][]{#1}%
\maketitle
\begin{center}
    \vspace{-0.5cm}
    \centering
    \includegraphics[width=\textwidth]{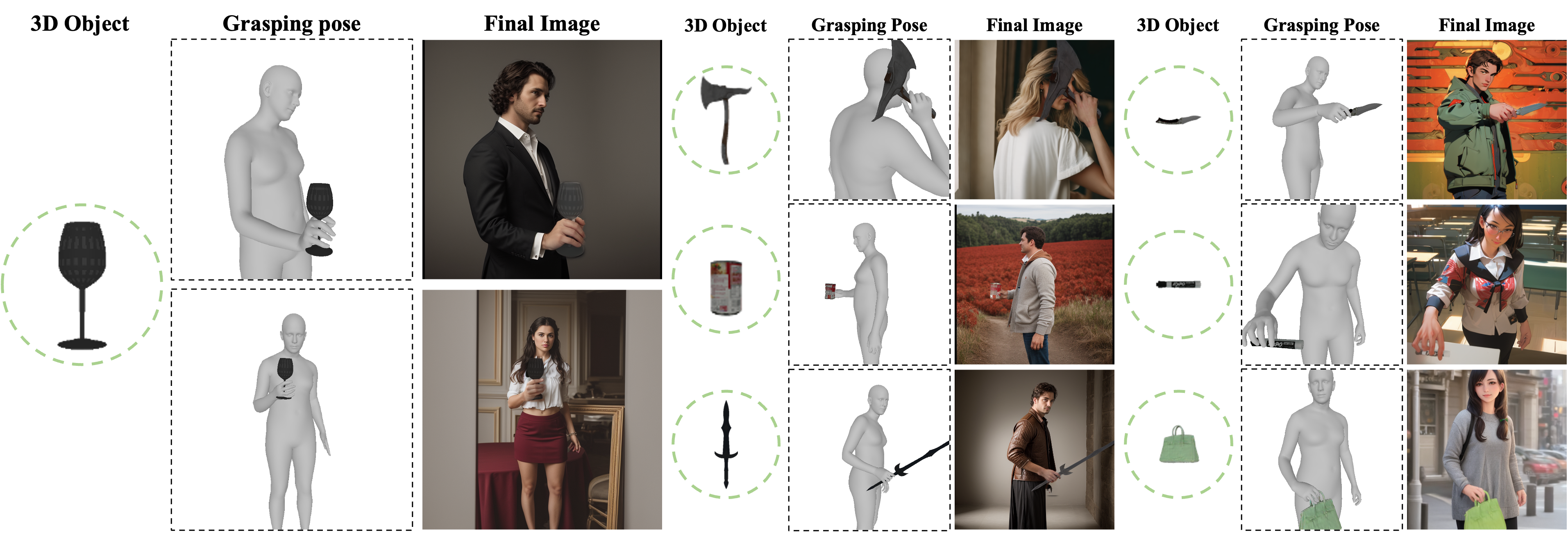}
    \captionof{figure}{Given an object mesh and its relative position, GraspDiffusion generates whole body grasping 3D poses, which is subsequently used as guidance for creating human-object interaction scenes. As shown, GraspDiffusion can synthesize images with valid human-object interactions for various types of objects. Note that the bottom-right sample (a green bag) was created from an object image, which was made into a 3D using TripoSR \cite{TripoSR2024}, further paving the way for various use cases.}
    \label{fig:teaser}
\end{center}
}]

\begingroup
\renewcommand{\thefootnote}{\fnsymbol{footnote}} 
\footnotetext[1]{Work done while at Naver Webtoon.}
\endgroup

\input{sec/0_abstract}    
\input{sec/1_intro}

\input{sec/2_related}
\input{sec/3_method}
\input{sec/4_experiment}
\input{sec/5_conclusion}

{
    \small
    \bibliographystyle{ieeenat_fullname}
    \bibliography{main}
}


\clearpage
\setcounter{section}{0}
\renewcommand{\thesection}{S\arabic{section}}

\twocolumn[
    \begin{center}
        \Large\bfseries
        Supplementary Material for GraspDiffusion: Synthesizing Realistic Whole-body Hand-Object Interaction
    \end{center}
    \vspace{1em}  
]

\input{sec/x_supplementary}

\end{document}

%% file: sec/0_abstract.tex
\vspace*{-1\baselineskip} 
\begin{abstract}
\vspace*{-1.2\baselineskip} 

   Recent generative models can synthesize high-quality images, but they often fail to generate humans interacting with objects using their hands. This arises mostly from the model's misunderstanding of such interactions and the hardships of synthesizing intricate regions of the body. In this paper, we propose \textbf{GraspDiffusion}, a novel generative method that creates realistic scenes of human-object interaction. Given a 3D object, GraspDiffusion constructs whole-body poses with control over the object's location relative to the human body, which is achieved by separately leveraging the generative priors for body and hand poses, optimizing them into a joint grasping pose. This pose guides the image synthesis to correctly reflect the intended interaction, creating realistic and diverse human-object interaction scenes. We demonstrate that GraspDiffusion can successfully tackle the relatively uninvestigated problem of generating full-bodied human-object interactions while outperforming previous methods. Our project page is available at \url{https://yj7082126.github.io/graspdiffusion/}
\vspace*{-1.3\baselineskip} 
\end{abstract}

%% file: sec/1_intro.tex
\section{Introduction}
\label{sec:intro}
The recent advent of diffusion-based generative models \cite{pmlr-v37-sohl-dickstein15, song2021scorebased, Ho2020DenoisingDP} has demonstrated significant success in producing high-quality visual content \cite{Rombach2021HighResolutionIS, pmlr-v162-nichol22a, Ramesh2022Dalle2, saharia2022photorealistic, podell2024sdxl}. When trained on large datasets, these models can coherently synthesize images of various subjects corresponding to given textual/visual cues. However, despite their strong performance, generative models struggle to comprehend and visualize everyday hand-object interactions. This limitation hinders their broader adoption in generative model-based content creation pipelines.

\begin{figure}
    \begin{tabular}{ccc}
        \begin{subfigure}[b]{0.3\columnwidth}
            \centering
            \subcaption{ControlNet}
            \includegraphics[width=\textwidth]{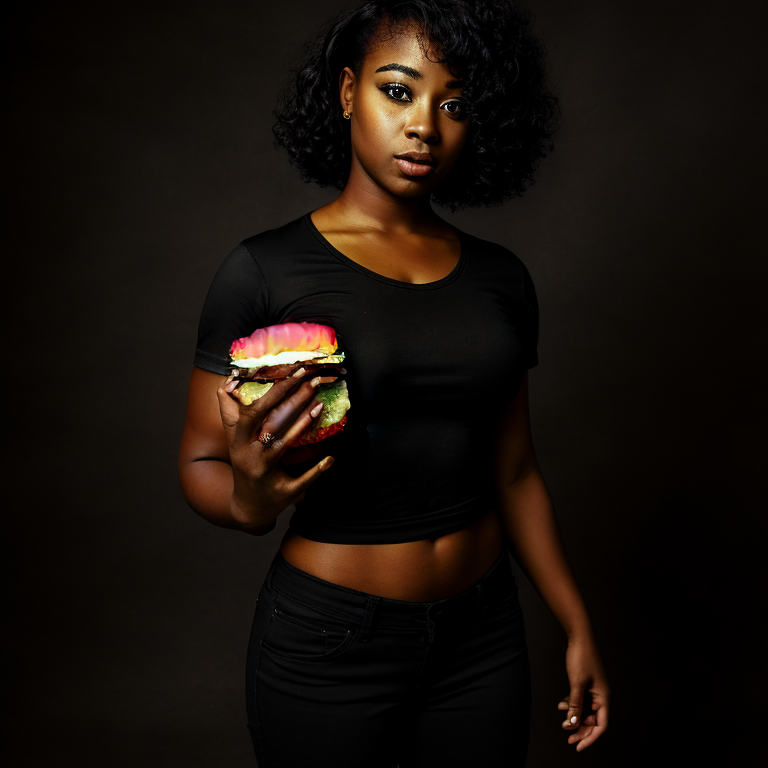}
        \end{subfigure} &
        \hfill
        \begin{subfigure}[b]{0.3\columnwidth}
            \centering
            \subcaption{HandRefiner}
            \includegraphics[width=\textwidth]{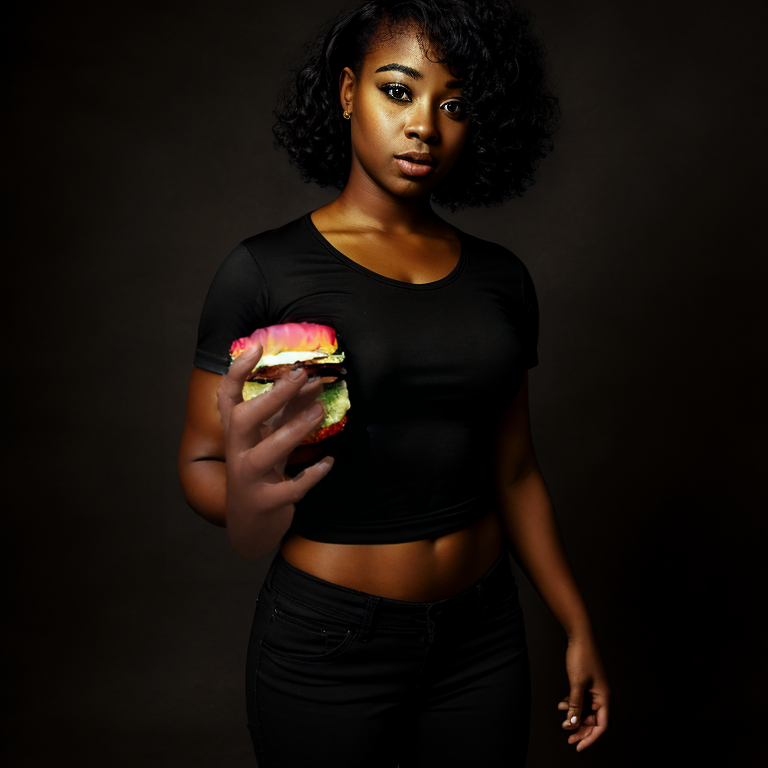}
        \end{subfigure} & 
        \hfill
        \begin{subfigure}[b]{0.3\columnwidth}
            \centering
            \subcaption{Ours} 
            \includegraphics[width=\textwidth]{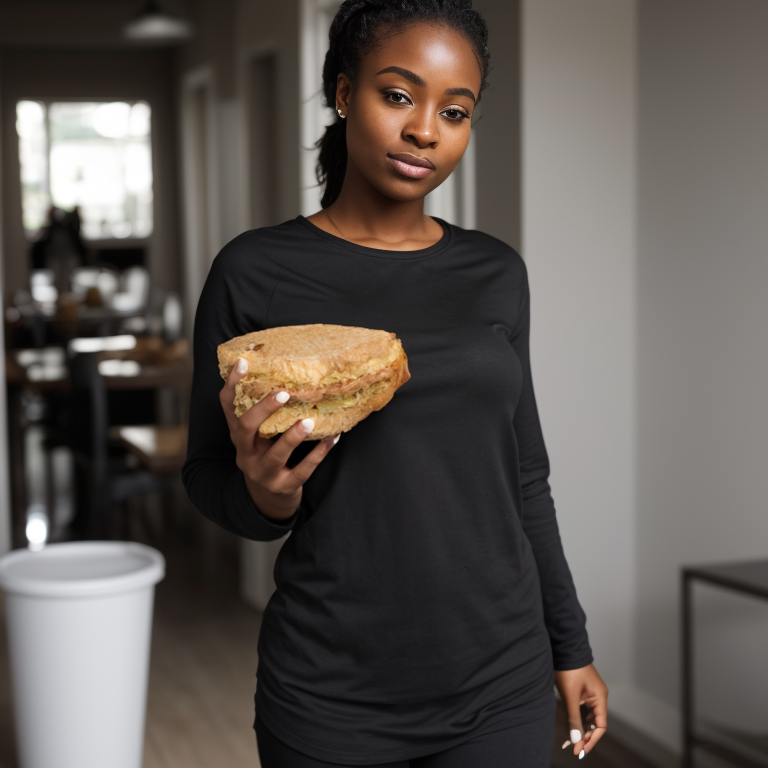}
        \end{subfigure} \\
        \begin{subfigure}[b]{0.3\columnwidth}
            \centering
            \includegraphics[width=\textwidth]{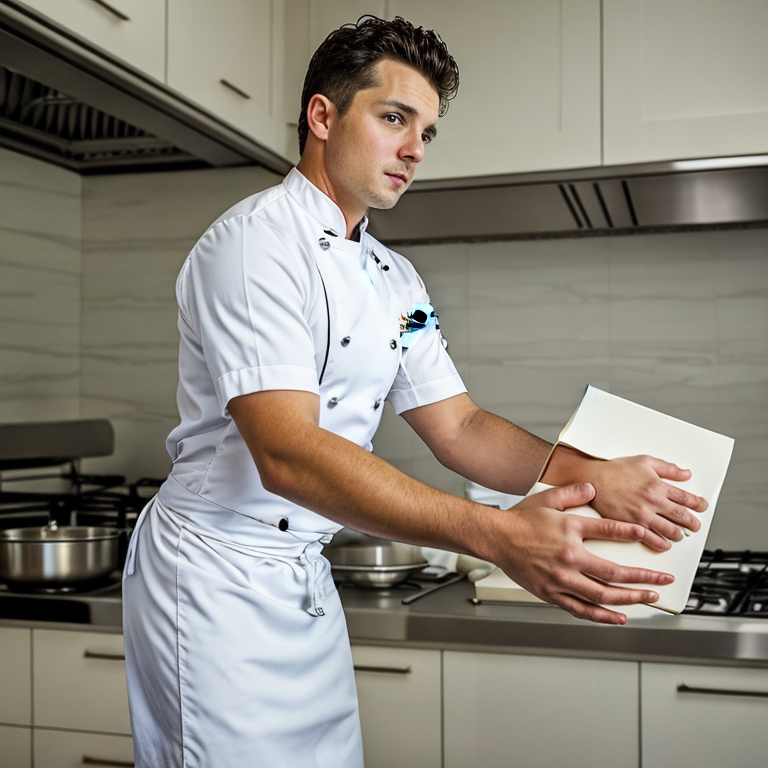}
        \end{subfigure} &
        \hfill
        \begin{subfigure}[b]{0.3\columnwidth}
            \centering
            \includegraphics[width=\textwidth]{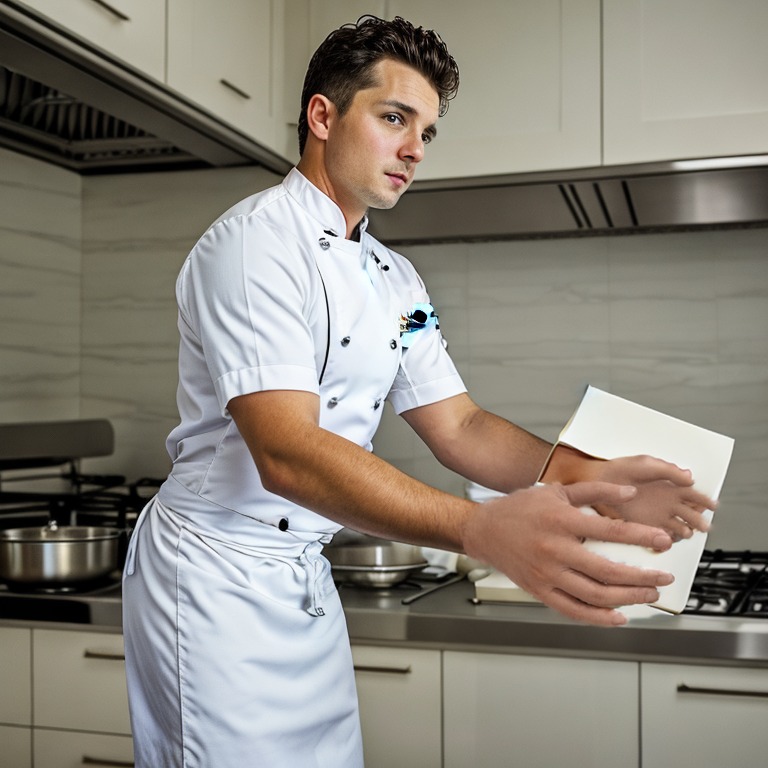}
        \end{subfigure} & 
        \hfill
        \begin{subfigure}[b]{0.3\columnwidth}
            \centering
            \includegraphics[width=\textwidth]{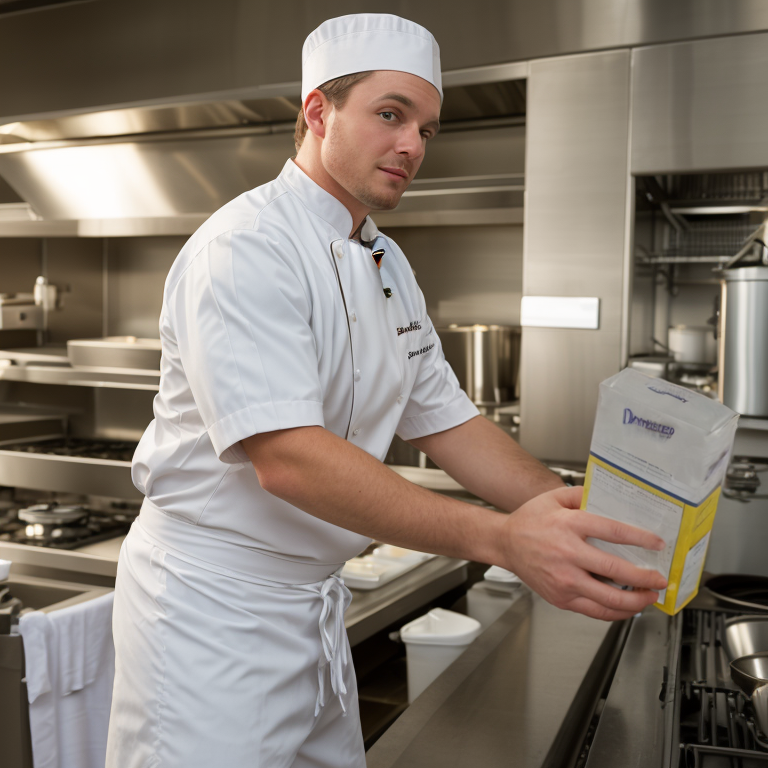}
        \end{subfigure} \\
        \begin{subfigure}[b]{0.3\columnwidth}
            \centering
            \includegraphics[width=\textwidth]{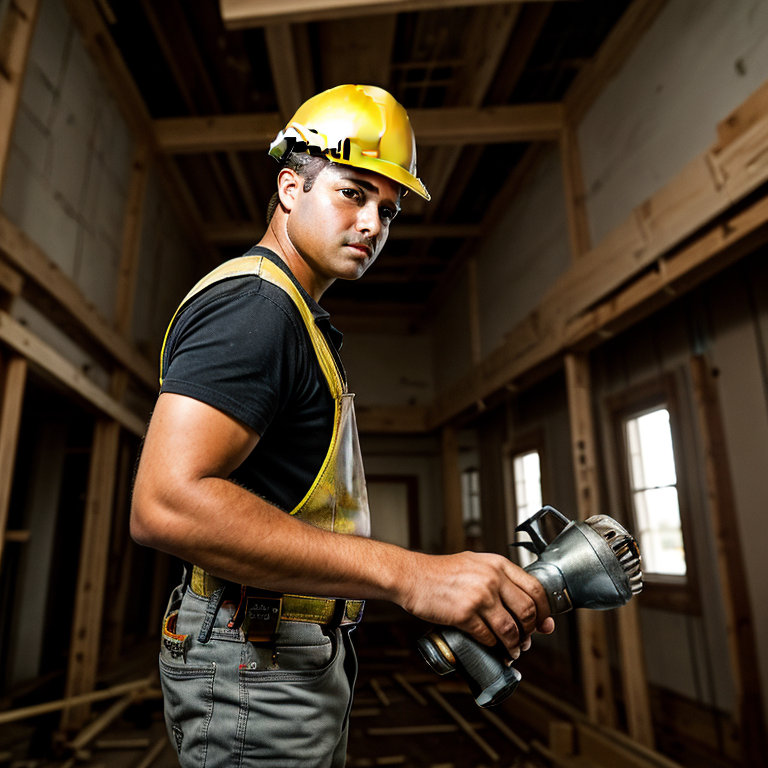}
        \end{subfigure} &
        \hfill
        \begin{subfigure}[b]{0.3\columnwidth}
            \centering
            \includegraphics[width=\textwidth]{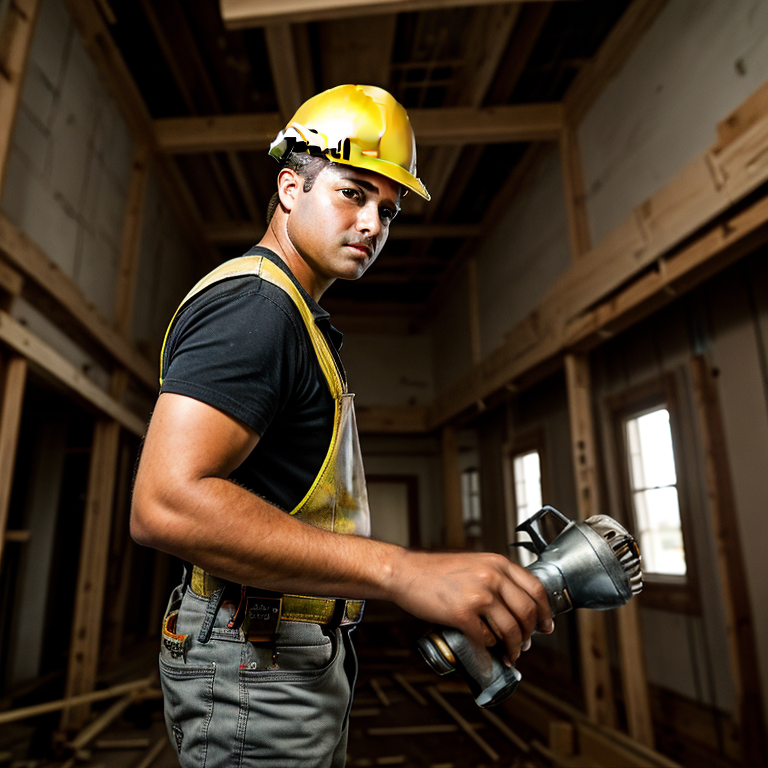}
        \end{subfigure} & 
        \hfill
        \begin{subfigure}[b]{0.3\columnwidth}
            \centering
            \includegraphics[width=\textwidth]{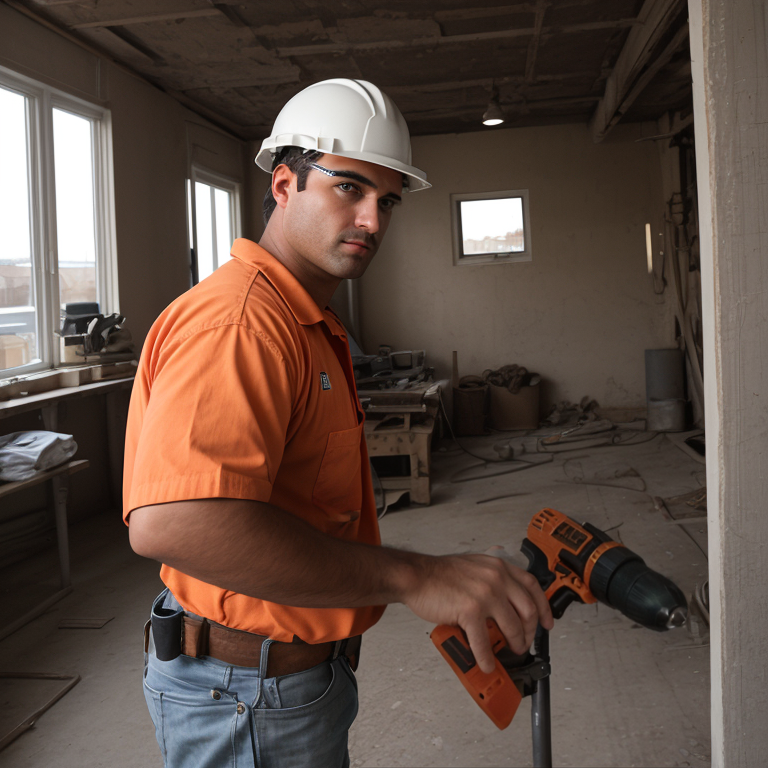}
        \end{subfigure} \\
   \end{tabular}
   \vspace{-0.25cm}
  \caption{Comparison between our method and previous approaches on generating HOI images. While previous methods can generate images conditioned on human pose and refine hand shapes, they are prone to erroneous object creation (top row) or faulty interaction synthesis (bottom row).}
  \label{fig:figure2_neg}
  \vspace{-0.5cm}
\end{figure}

One challenge in generating images with high-quality hands arises from the fact that hands occupy only marginal areas within a full-bodied human image, yet have a complex anatomical structure that presents a wide variety of possible hand poses. Hands come in varying shapes, sizes, orientations, and multiple finger joints that can bend in various degree to support diverse hand poses. Moreover, hands are usually interacting with various objects, making the distribution of hands highly complex and convoluted, leading to faulty generation results (e.g., distorted hand poses, multiple arms from a shoulder, uncanny hand shapes). Examples of such inaccurate generation are displayed in Fig.~\ref{fig:figure2_neg}. 

While several papers \cite{lu2023handrefiner, wang2024rhandsrefiningmalformedhands, Pelykh2024GivingAH} have applied inpainting for hand region refinement, they only focus on situations where the hand is not interacting with other objects, making it impractical for most use cases. In order to generate realistic grasping hands for a given object, the models must understand the semantics and functionalities provided by the object--a concept well known as affordance \cite{Gibson1979TheEA}. While Affordance Diffusion \cite{Ye2023AffordanceDS} and HOIDiffusion \cite{zhang2024hoidiffusion} creates an image of a single grasping hand using affordance, these methods only represent the explicit physical contact between a human hand and the object (devoid of any human identity), and fail to convey the spatial / orientational non-contact relationships, making them unsuitable for understanding human affordance. For instance, when using a cell phone, the relationships between the human's face and torso with respect to the phone should be considered as part of interaction along with the hand touching the phone. This requirement necessitates the development of an pipeline that creates identifiable human-object images where the human's body is visible, such that the implicit, non-contact relationships are well captured.

In this paper, we present GraspDiffusion, a novel method for generating interaction images with realistic hands and a clear human identity from a single 3D object input. Instead of relying on textual prompts, which tend to be ambiguous in describing complex interactions \cite{Narasimhaswamy_2024_CVPR}, we first utilize a diffusion model to generate full-bodied grasping poses \cite{SMPL-X:2019, MANO:SIGGRAPHASIA:2017} conditioned on the object and its position. The generated 3D pose parameters are provided as conditions for the next stage to create high-quality image samples. Specifically, we train multiple conditional models \cite{Zhang_2023_ICCV, Mou2023T2IAdapterLA} and a novel cross-attention modulation scheme \cite{balaji2022eDiff-I, Couairon2023ZeroshotSL} to correctly convey the interaction context without harming the diversity of the generated samples. To overcome the lack of a large-scale image-3D pose paired dataset for human object interaction, we also propose a dataset annotation scheme to gather 3D annotations from previous image-based interaction datasets \cite{Gupta2015VCOCO, Gkioxari2017DetectingAR, Chao2017HICODET, Li2019HAKEHA, Jin2020WholeBodyHP}. 

As far as our knowledge, our pipeline is the first approach to generate full-bodied HOI images from a given object information, such that both explicit and implicit interactions are portrayed in a physically plausible manner. Our experiments on several metrics show that GraspDiffusion outperforms similar approaches in generating high-quality images with realistic interactions and valid 3D grasping poses. We also display cases where GraspDiffusion can be conditioned with a single object image and support various artistic styles, proving its efficacy as a practical solution for AI practitioners in content creation. The contributions of this work are summarized as follows.




\begin{itemize}[leftmargin=*]
\item We propose GraspDiffusion, a novel generative pipeline that synthesizes realistic, full-bodied HOI images.
\item We divide the pipeline into two stages to facilitate both physically plausible body poses along with identity and style diversity. The first stage provides a rich 3D prior to create lifelike interaction poses, which are then provided as 3D conditions for the image generation stage which prodces images with high quality and diversity.
\item We devise an annotation pipeline for an image-3D paired grasping dataset to facilitate the training of HOI image generation. 
\item Our experiments in generating realistic images and their paired 3D grasping poses demonstrate the efficacy and substantial performance improvements over previous SOTA methods, providing advanced physical pose plausibility and perceptive visual quality.
\end{itemize}



\begin{figure*}
  \centering
  \includegraphics[width=\textwidth]{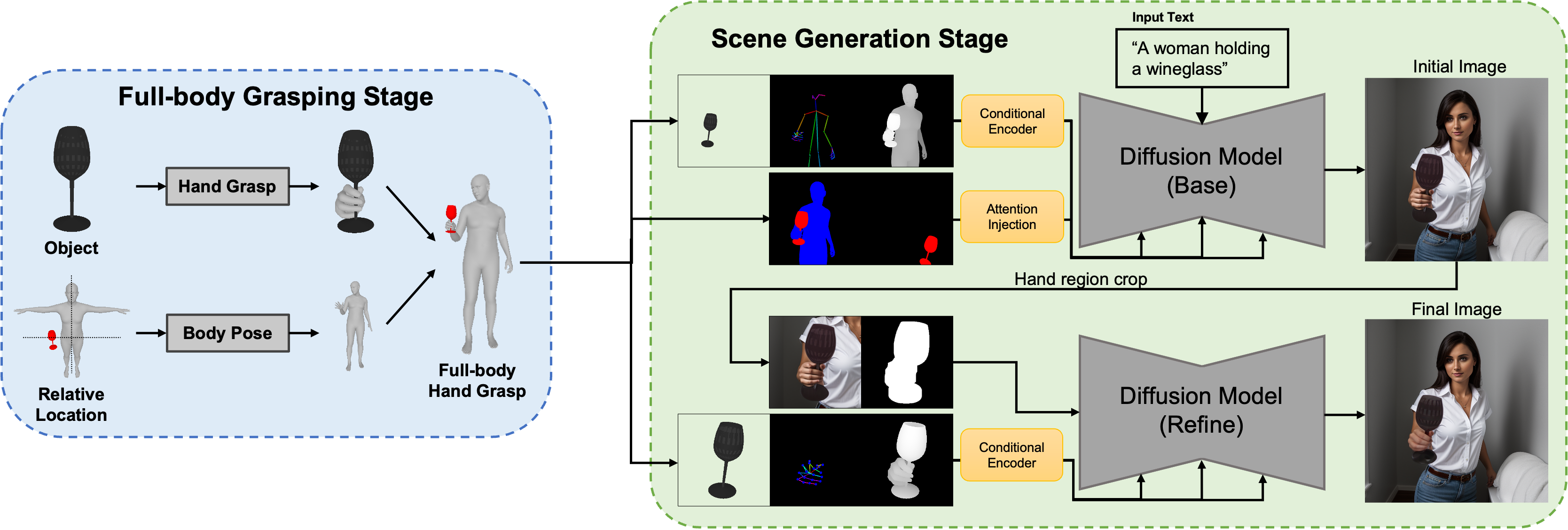}
  \caption{We present a two-stage pipeline to generate realistic human-object-interaction images. The first stage takes a single object model and its human-centric location to synthesize a 3D full-bodied grasping pose, providing scene-level context for image generation. The second stage takes reference from the 3D grasping pose, conditionally generating high-quality images.}
  \label{fig:main_arch}
  \vspace{-0.25cm}
\end{figure*}

%% file: sec/2_related.tex
\section{Related Work}
\label{sec:related}

\textbf{Conditional Image Generation.} To provide additional fine-grained, spatial conditions for diffusion models, ControlNet \cite{Zhang_2023_ICCV} and T2I-Adapter \cite{Mou2023T2IAdapterLA} proposed using image-level signals to control the generation process, which includes using 2D human keypoint skeletons \cite{Cao2018OpenPoseRM, Yang2023EffectiveWP} for human pose guidance and depth maps for better depth perception. While the improvements in conditional image generation have been significant, synthesizing humans interacting with objects hasn't reached the same level of improvement. 

Several papers \cite{lu2023handrefiner, wang2024rhandsrefiningmalformedhands, Pelykh2024GivingAH} proposed to refine the hands of images generated using Controlnet-based inpainting, but does not count as a direct solution to human-object interaction. Others focused on identifying contacts for inpainting a new hand or object for a given scene \cite{Ye2023AffordanceDS, Kulal2023PuttingPI, Min2024GenHeldGA, xue2024hoiswap}, yet they are limited in camera views and require a prior scene context, making them infeasible in practical scenarios. HOIDiffusion \cite{zhang2024hoidiffusion} also generates images from synthesized grasps, but is limited to hand-centric views. Compared to HanDiffuser \cite{Narasimhaswamy_2024_CVPR}, which applied the injection of hand embeddings during image generation to create realistic hands, our pipeline focuses more on the joint synthesis of hand and object, and uses spatial guidance to better direct the generation process towards a 3D scene context.

\textbf{Grasp Synthesis.} Synthesizing a hand grasp consisting of a given object and a hand model is important in understanding human-object interaction, and is a widely studied task in robotics, graphics, and computer vision. 
While the focus in robotics is to make stable grasps for a given object in simulation / real life \cite{Li2022GenDexGraspGD, Urain2022SE3DiffusionFieldsLS, Wang2022DexGraspNetAL, Geng2023UniDexGraspID, Weng2024DexDiffuserGD}, in computer vision and graphics the focus is to make plausible grasps that are physically plausible, and generate grasps for either hands \cite{Christen2021DGraspPP, Jiang2021GraspTTA, Karunratanakul2020GraspingFL, Zhou2022TOCHSO, Zheng2023CAMSCM, Christen2024DiffH2ODS, Liu2024GeneOHDT, Li2024ClickDiffCT} or a human \cite{Tendulkar2022FLEXFG, Ghosh2022IMoSIF, Taheri2021GOALG4, Wu2021SAGASW, Braun2023PhysicallyPF, Zheng_2023_ICCV, Luo2024GraspingDiverse}. 

Thanks to the advent of human-object interaction datasets \cite{Bhatnagar2022BEHAVEDA, Brahmbhatt_2020_ECCV, Chao2021DexYCBAB, Fan2022ARCTICAD, Guzov2022VisuallyPH, Hampali2020honnotate, Kwon2021H2OTH, Liu2022HOI4DA4, Taheri2020GRABAD}, many grasp synthesis methods achieved high performance in generating plausible grasps. Yet most existing datasets have issues with data scalability and variability, especially when it comes to color image-3D paired datasets. While hand-object interaction datasets like DexYCB \cite{Chao2021DexYCBAB}, ARCTIC \cite{Fan2022ARCTICAD} and HOI4D \cite{Liu2022HOI4DA4} consist of richly annotated image data, they are rather focused only on the hand and object recorded in an egocentric manner, devoid of any human identity and spatial / orientational non-contact relationships. Although BEHAVE \cite{Bhatnagar2022BEHAVEDA} contains both RGB video sequences with 3D annotation, the image quality and motion sensors worn by the subject make it difficult to use as a realistic image dataset. To overcome this issue, we used traditional human-object interaction datasets \cite{Gupta2015VCOCO, Gkioxari2017DetectingAR, Chao2017HICODET, Li2019HAKEHA, Jin2020WholeBodyHP} along with annotation tools \cite{Kirillov2023SegmentA, liu2023grounding, li2021hybrik, li2023hybrikx, ke2023marigold} to create pseudo-3D annotations for the 2D image.

Building our insight from similar approaches, \cite{Tendulkar2022FLEXFG, Zheng_2023_ICCV}, we focus on leveraging priors from a full-body pose model and a hand-grasping model. Compared to previous approaches \cite{Taheri2021GOALG4, Ghosh2022IMoSIF, Wu2021SAGASW, Braun2023PhysicallyPF}, instead of generating a motion sequence, we synthesize the pose parameters for the 3D parametric models \cite{SMPL-X:2019, MANO:SIGGRAPHASIA:2017} using a diffusion model.

%% file: sec/3_method.tex
\section{GraspDiffusion}

Fig. \ref{fig:main_arch} illustrates the proposed architecture. Starting with a 3D object mesh and its position within the human-centric coordinate system (originating at the pelvis joint), GraspDiffusion synthesizes realistic images portraying a human interacting with the object, with a significant portion of the human body visible to be considered "full-body". In the initial full-body grasping stage, we generate the pose parameters for the human body model \cite{SMPL-X:2019} interacting with the 3D object mesh (Section.~\ref{subsection_1}). In the scene generation stage, we extract geometric structures from the pose parameters to guide the generation of realistic images, leveraging a latent diffusion model \cite{Rombach2021HighResolutionIS} along with spatial encoders and a cross-attention modulation scheme (Section.~\ref{subsection_2}).

\subsection{Preliminaries}

\textbf{Diffusion Models.} Diffusion models \cite{pmlr-v37-sohl-dickstein15, Ho2020DenoisingDP} are a group of generative models that interpret the data distribution $p(x)$ as a sequential transformation from a tractable prior distribution $p(x_T) \sim \mathcal{N}(0, I)$. During training, the model uses a forward noise process $q(x_t|x_{t-1})$ that gradually adds a small amount of noise to a clean data sample $x_0$ towards $p(x_T)$. At the same time, the model learns a backward noise process $p(x_{t-1}|x_t)$ implemented as a neural network, which is trained to remove the noise from the before generating samples from $p(x_T)$. For latent diffusion models \cite{Rombach2021HighResolutionIS}, the diffusion process is performed in the latent space of a trained autoencoder model \cite{Kingma2013VAE}, guided by a conditional text embedding derived from the CLIP \cite{Radford2021CLIP} mechanism. 


\textbf{3D Parametric Models.} For the hand model, we use the MANO \cite{MANO:SIGGRAPHASIA:2017} differentiable model, in which we input the full finger articulated hand pose $\theta_{\text{h}} \in \mathbb{R}^{15 \times 3}$, wrist translation $t_{\text{h}} \in \mathbb{R}^{3}$ and global orientation $R_{\text{h}} \in \mathbb{R}^{3}$ and get a 3D mesh $\mathcal{M}_{\text{h}}$ with vertices $\mathcal{V}_{\text{h}}$. For the full-body model, we use the SMPL-X \cite{SMPL-X:2019} differentiable model, in which we input the full-body pose $\theta_{\text{b}} \in \mathbb{R}^{21 \times 3}$, the full finger articulated hand pose $\theta_{\text{h}} \in \mathbb{R}^{15 \times 3}$ for both hands, the root translation $t_{\text{b}} \in \mathbb{R}^{3}$ and global orientation $R_{\text{b}} \in \mathbb{R}^{3}$ and get a 3D mesh $\mathcal{M}_{\text{body}}$ with vertices $\mathcal{V}_{\text{b}}$. 


\begin{figure}
  \centering
  \includegraphics[width=0.48\textwidth]{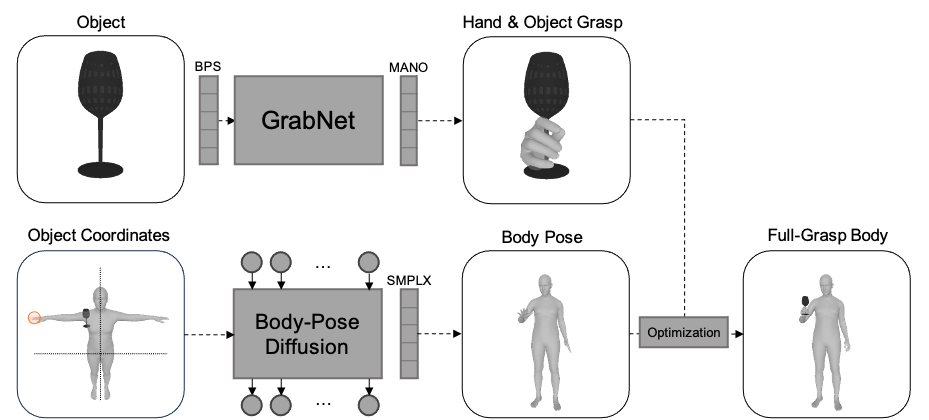}
  \captionof{figure}{Full-body grasping pipeline. We separately leverage a hand-grasping model \cite{Taheri2020GRABAD} and a body-pose diffusion model, and perform a joint optimization into a full-bodied grasping pose.}
  \label{fig:3d_arch}
  \vspace{-0.5cm}
\end{figure}

\subsection{Full-Body Grasping Pipeline} \label{subsection_1}

 
Building on prior approaches \cite{Tendulkar2022FLEXFG, Zheng_2023_ICCV}, we separately generate hand grasps and body poses in creating a whole-body grasping pose. Specifically, we take a 3D object mesh, its relative location to the human root, and the contacting hand orientation (left or right) as the input, to generate an SMPL-X mesh that grasps the given 3D object with a realistic body pose and hand-object contact. 

The input object mesh is used to generate a plausible MANO \cite{MANO:SIGGRAPHASIA:2017} hand grasp, for which we utilized GrabNet \cite{Taheri2020GRABAD}, a conditional variational autoencoder (cVAE) that produces hand grasps conditioned on the Basis Point Set (BPS) \cite{Prokudin2019EfficientLO} of the given object. Separately trained for left and right-hand grasps, GrabNet generates MANO parameters $(\theta_\text{h}, t_\text{h}, R_\text{h})$ that grasps the given object, displaying accurate contact and high generalization for unseen objects. 

The object's relative location $t_\text{obj} \in \mathbb{R}^3$ and the hand orientation $c_\text{left}, c_\text{right} \in \{0,1\} $ is then used to create a body pose that not only roughly positions its hand in the desired object location, but also reflects the appropriate implicit relationships required for a plausible grasping body pose \cite{Taheri2021GOALG4, Kim2024BeyondTC}; whether the head is correctly oriented towards the object, the arms are correctly extended and the torso is leaning towards the object. To achieve this, we utilize a diffusion generative model trained to generate SMPL-X pose parameters $(\theta_\text{body}, R_\text{body})$ conditioned on an object location and whether to use the right/left hand for contact. The loss is defined as
\begin{equation}
\mathcal{L}_{DM} = \mathbb{E}_{x, \epsilon \sim \mathcal{N}(0,I),t} [ || \epsilon - \epsilon_\theta (x_t, t, c)^2_2 || ],
\label{eq:1}
\end{equation}
where $c = [t_\text{obj}, c_\text{left}, c_\text{right} ] \in \mathbb{R}^5$ and $x \in \mathbb{R}^{132}$, which consists of the 6 DoF global orientation and body pose. 

We then apply the finger articulation of the hand grasp to the body pose, creating an initial full-body grasping pose. To correctly align the 3D hand-object grasp with the human body, we optimize over the rotation $(R_\text{h})$ and translation $(t_\text{h})$ of the MANO hand model while retaining the original finger articulation. Focusing on the palm region of the hands, given vertices $\mathcal{V}_{h}^{p}$ (output palm vertices from MANO) and corresponding vertices $\mathcal{V}_{b}^{p}$ (output palm vertices from SMPL-X), we align them using:
\begin{equation}
E(R_\text{h}, t_\text{h}) = \frac{1}{|\mathcal{V}_{h}^{p}|} \sum_{i=1}^{|\mathcal{V}_{h}^{p}|} d_\text{vv} (\mathcal{V}_{h_i}^{p}, \mathcal{V}_{b_i}^{p}),
\label{eq:3}
\end{equation}
where $d_\text{vv}$ represents the $L^1$ distance between the two vertices in the 3D space. The optimized $(R_\text{h}, t_\text{h})$ is used to transform the 3D object, correctly positioning it within the full-body grasping pose as it was for the hand-object grasp, completing the grasping pose.

\begin{figure}
  \centering
  \includegraphics[width=0.4\textwidth]{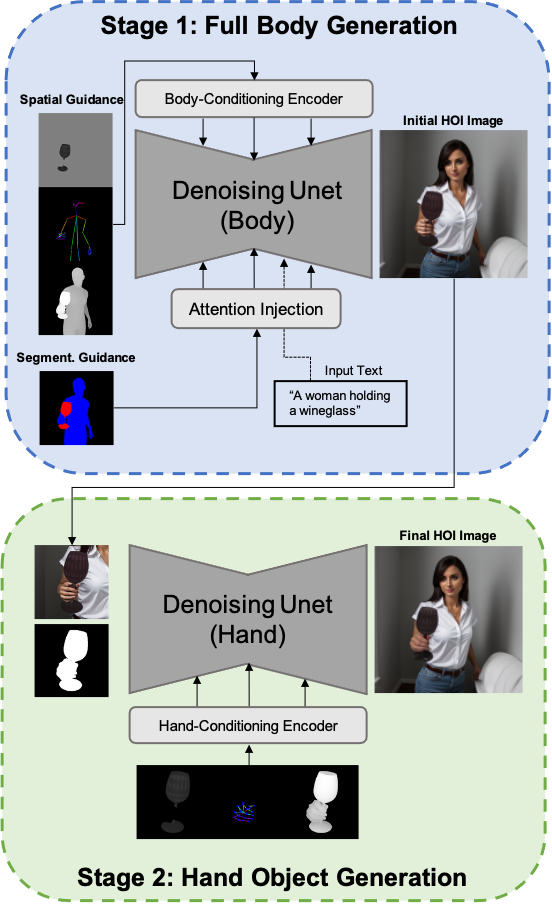}
  \captionof{figure}{Scene generation stage. We inject three image conditions and semantic segmentation images as guidance for the generation of a high-quality HOI image. We then use the same types of renderings centered on the hand-object region to refine the hand quality.}
  \label{fig:2d_arch}
\end{figure}

\subsection{Scene Generation Pipeline} \label{subsection_2}

Given the 3D body pose pose, the scene generation pipeline extracts multiple spatial conditions as conditions to a pre-trained \cite{Rombach2021HighResolutionIS} model to create consistent images of human-object interaction. Optionally, the pipeline can refine the image focused on the hand-object region to further adjust interaction and correct erroneous details.


\begin{figure}
  \centering
  \includegraphics[width=0.45\textwidth]{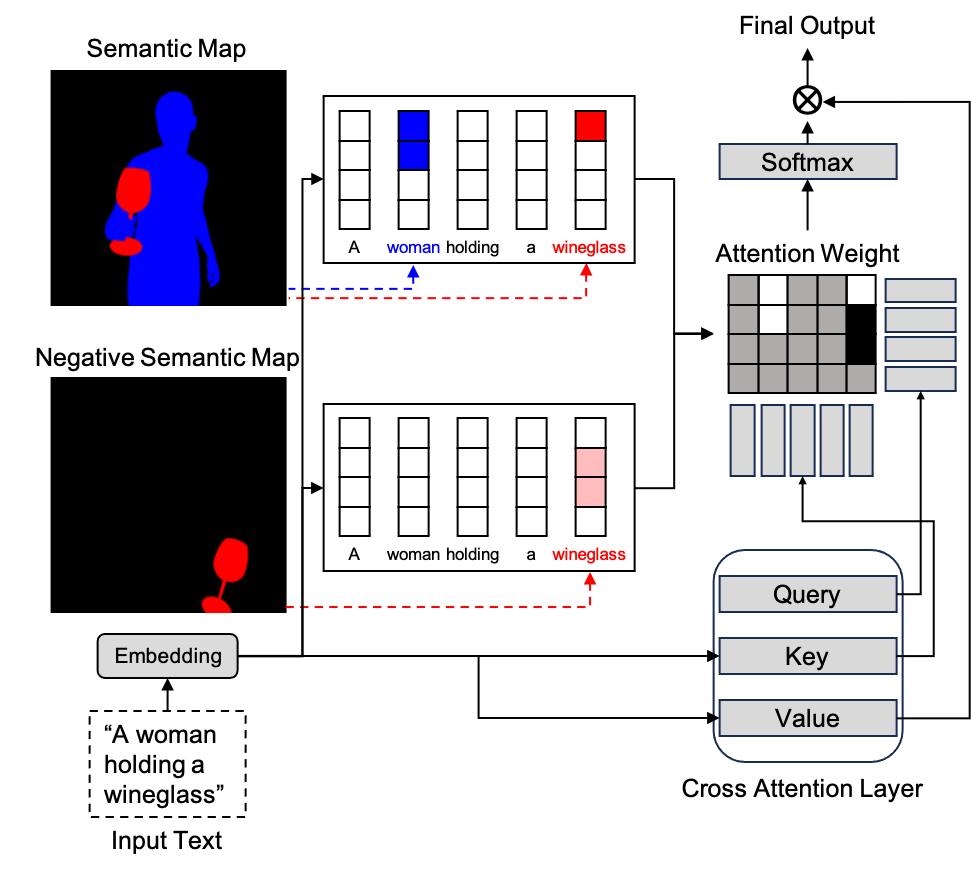}
  \captionof{figure}{Attention Injection Scheme. During inference, we inject the human/object semantic maps into the cross-attention layers as guidance, encouraging the generation process to be focused on the segmented regions. We also apply a negative semantic map for the object to avoid undesired cases where the opposite hand interacts with the object.}
  \label{fig:pww}
  \vspace{-0.5cm}
\end{figure}

To precisely control the human-object image's generation, we render three spatial conditions from the full-body grasping output. We first render the skeleton $(s^i)$ of the SMPL-X body, consisting of body and hand joints, to ensure realistic human proportions within the generated image. We also use the joint depth map $(d^i)$ from the SMPL-X and object model to provide depth information. Lastly, we render the occluded object with ambient lighting $(o^i)$ to preserve its appearance while relighting it. To apply conditions, we chose the CoAdapter \cite{Mou2023T2IAdapterLA} approach, which allows flexibility in handling multiple conditions. For each condition, we separately apply an adapter $\mathcal{F}_\text{AD}$ and perform a weighted sum to create feature $\mathbf{F}_{c}$, which can be written as
\begin{equation}
\mathbf{F}_{c} = \sum_{k \in \{ s, d, o \}} \omega_{k}\mathcal{F}_\text{AD}^{k}(k^i).
\label{eq:4}
\vspace{-0.2cm}
\end{equation}
During training of the conditioning pipeline, we fix the parameters of the baseline U-Net model and only optimize the conditional adapters, reducing the risk of the model converging to the dataset's style. This decision allows us to control the image style by applying LoRAs \cite{hu2022lora} or using finetuned Stable Diffusion models during inference time, making it suitable for real-world application tasks requiring personalized image generation.
\begin{equation}
\mathcal{L}_{ADM} = \mathbb{E}_{z, \epsilon \sim \mathcal{N}(0,I),t, \mathbf{F}_c} [ || \epsilon - \epsilon_\theta (z_t, t, c_\text{text}, \mathbf{F}_c)^2_2 || ].
\label{eq:2}
\end{equation}

For hand-object refinement, we utilize the full-body grasping output to produce a joint hand-object mask and spatial conditions: hand skeleton information $(s_h^i)$, a joint hand-object depth map $(d_h^i)$, and the occluded rendered object $(o_h^i)$. These masks and conditions serve as inputs to the hand refinement adapters, which is akin in structure to the body conditioning adapters yet trained separately. This refines the structure and appearance of both hand and object while preserving visual integrity. A full illustration of the procedure is shown in Fig.~\ref{fig:2d_arch}.


To address the issue of erroneous interactions, in which interactions may occur from locations other than the intended area (examples on Fig.~\ref{fig:comp1}), we introduce a training-free guidance method motivated from prior zero-shot semantic image synthesis techniques ~\cite{balaji2022eDiff-I, Couairon2023ZeroshotSL}. We first render binary segmentation masks from the posed human and object 3D model $(m^i, m_o^i)$, which are then sent to the cross-attention layers as guidance, down-sampled to match the resolution of each layer. Specifically, we create an input attention matrix $A \in \mathbb{R}^{{N_i} \times {N_t}}$ from the masks, applied to the cross-attention layers to encourage attention towards the intended region. We also modify the original procedure through the usage of a negative mask; specifically, we create a pseudo object segmentation map $m_{no}^i$ which, instead of using the intended hand, is using the opposite hand to grasp the 3D object model. This segmentation mask is then subtracted from the input attention matrix, disencouraging the generation in unintended locations.

%% file: sec/4_experiment.tex
\section{Experiments}

\begin{figure}[t]
    \centering
    \begin{subfigure}[b]{0.49\columnwidth}
        \centering
        \subcaption{Original Image}
        \includegraphics[width=\textwidth]{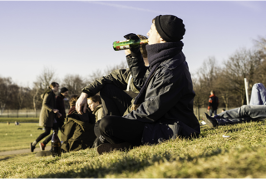}
    \end{subfigure}
    \hfill
    \begin{subfigure}[b]{0.49\columnwidth}
        \centering
        \subcaption{Joint depth map}
        \includegraphics[width=\textwidth]{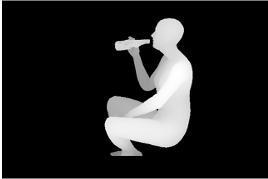}
    \end{subfigure}
    \begin{subfigure}[b]{0.49\columnwidth}
        \centering
        \subcaption{Human skeleton + Object}
        \includegraphics[width=\textwidth]{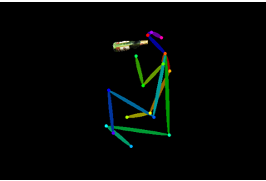}
    \end{subfigure}
    \hfill
    \begin{subfigure}[b]{0.49\columnwidth}
        \centering
        \subcaption{Segmentation map}
        \includegraphics[width=\textwidth]{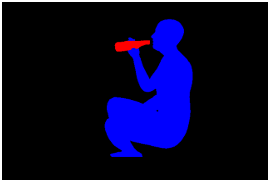}
    \end{subfigure}
  \caption{Processed HICO-DET \cite{Chao2017HICODET} dataset sample. We jointly extract the joint depth map for the object and human 3D pose, the human skeleton joint map, and semantic segmentation maps.}
  \vspace{-0.25cm}
  \label{fig:hico_sample}
\end{figure}

\subsection{Dataset Construction} To compensate for the lack of realistic, 3D-annotated human-object interaction datasets, we designed an annotation pipeline through which we leveraged previous interaction datasets \cite{Gupta2015VCOCO, Gkioxari2017DetectingAR, Chao2017HICODET, Li2019HAKEHA, Jin2020WholeBodyHP} to construct a pseudo-3D interaction dataset. Specifically, we utilized the human-object interaction images from HICO-DET \cite{Chao2017HICODET} and V-COCO \cite{Gupta2015VCOCO}, which contain a large variety of possible interactions and annotations for the human body and object type.

For both datasets, we first filtered the images so that each image included at least one visible human with a reasonable screen size, along with at least one identifiable hand. For datasets like HICO-DET, multiple humans within an image might be interacting with multiple objects, making it necessary to correctly identify the main object of interaction. We use the BLIP-2 language model \cite{Li2023BLIP2BL} to perform a Visual Question Answering task, which outputs an object type from the input image. Using the object text, we use GroundingDINO \cite{liu2023grounding} to get the object location and detect the segmentation map of the object, and employ a state-of-the-art depth estimation model \cite{ke2023marigold} to estimate the depth map of the object. Note that for V-COCO, we use the original annotated information for objects Meanwhile, we also estimate the 3D SMPL-X parameters for the human, using the annotated bounding box and HybrIK \cite{li2021hybrik, li2023hybrikx} 3D human pose and shape estimator. For the human skeleton, we use the annotated Halpe dataset \cite{Jin2020WholeBodyHP} for the HICO-DET dataset and the DWPose estimator \cite{Cao2018OpenPoseRM} for the V-COCO dataset. Among the processed images, we identify images with sufficiently large hands portrayed and reserve them for the hand-object refinement model training. To estimate the 3D MANO parameters, we use the ACR \cite{Yu2023ACRAC} hand pose and shape estimator. To further augment the dataset, we use the BEHAVE interaction dataset \cite{Bhatnagar2022BEHAVEDA} that comes with SMPL-X parameters and object 3D models, and processed it in a similar manner to create joint image interaction pairs (image, text, depth map, skeleton, segmentation).

For the hand refinement modules, we process the Dex-YCB dataset \cite{Chao2021DexYCBAB}, the RHD dataset \cite{Zimmermann_2017_ICCV} that also comes with 3D annotations and video data. By processing the MANO parameters \cite{MANO:SIGGRAPHASIA:2017} and the object 3D models, we also acquired image interaction pairs specifically focused on the hand region. To broaden the hand-object interaction distribution, we include a subset of the preprocessed HICO-DET dataset, cropped on the hand-object bounding box, as additional training data for hand refinement. In total, we collected 25K joint interaction pairs of the full human-object interaction scenes to train the scene generation pipeline, and 15K joint interaction pairs of the hand-object interaction scenes to train the hand refinement pipeline.



\begin{table}
    \small
    \caption{Quantitative comparison on full-bodied generation}
    \vspace{-0.25cm}
    \centering
    \resizebox{\columnwidth}{!}{%
        \begin{tabular}{lccc}
            \toprule
            Methods & FID $\downarrow$ & KID $\downarrow$ & CLIPScore $\uparrow$  \\
            \midrule
            LDM (finetuned) \cite{Rombach2021HighResolutionIS} & 41.23 & $1.45 \times 10^{-2}$ & 0.671 \\
            ControlNet \cite{Zhang_2023_ICCV} & 32.76 & $1.23 \times 10^{-2}$ & 0.71 \\
            Champ \cite{zhu2024champ} & 40.63 & $2.23 \times 10^{-2}$ & 0.739 \\
            \midrule
            Ours (w/o atttention) & \textbf{22.55} & $5.63 \times 10^{-3}$ & 0.717 \\
            Ours & 22.88 & $\mathbf{5.55 \times 10^{-3}}$ & \textbf{0.767} \\
            \bottomrule
        \end{tabular}
    }
    \label{tab:comparisonfid}
    \vspace{-0.25cm}
\end{table}

\begin{table}
    \small
    \caption{Quantitative comparison on hand-object generation}
    \vspace{-0.25cm}
    \centering
    \resizebox{\columnwidth}{!}{%
        \begin{tabular}{lcccc}
        \toprule
        Methods & FID $\downarrow$ & KID $\downarrow$ & Hand Contact $\uparrow$ \\
        \midrule
        ControlNet \cite{Zhang_2023_ICCV} & 99.38 & $7.70 \times 10^{-2}$ & 58.17 \\
        HandRefiner \cite{lu2023handrefiner} & 92.48 & $7.11 \times 10^{-2}$ & 61.45 \\
        Affordance Diffusion \cite{Ye2023AffordanceDS} & - & - & 65.69 \\
        \midrule
        Ours & \textbf{64.67} & $\mathbf{4.36 \times 10^{-2}}$ & \textbf{97.94} \\
        \bottomrule
        \end{tabular}
    }
    \label{tab:comparisonhand}
    \vspace{-0.25cm}
\end{table}

\begin{figure*}
    \vspace{-0.25cm}
    \resizebox{\textwidth}{!}{%
    \begin{tabular}{cccccc}
        \begin{subfigure}[b]{0.15\textwidth}
            \centering
            \subcaption{Input Object}
            \includegraphics[width=\linewidth]{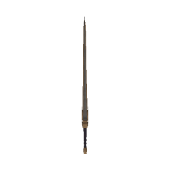}
        \end{subfigure} &
        \begin{subfigure}[b]{0.15\textwidth}
            \centering
            \subcaption{SDXL}
            \includegraphics[width=\linewidth]{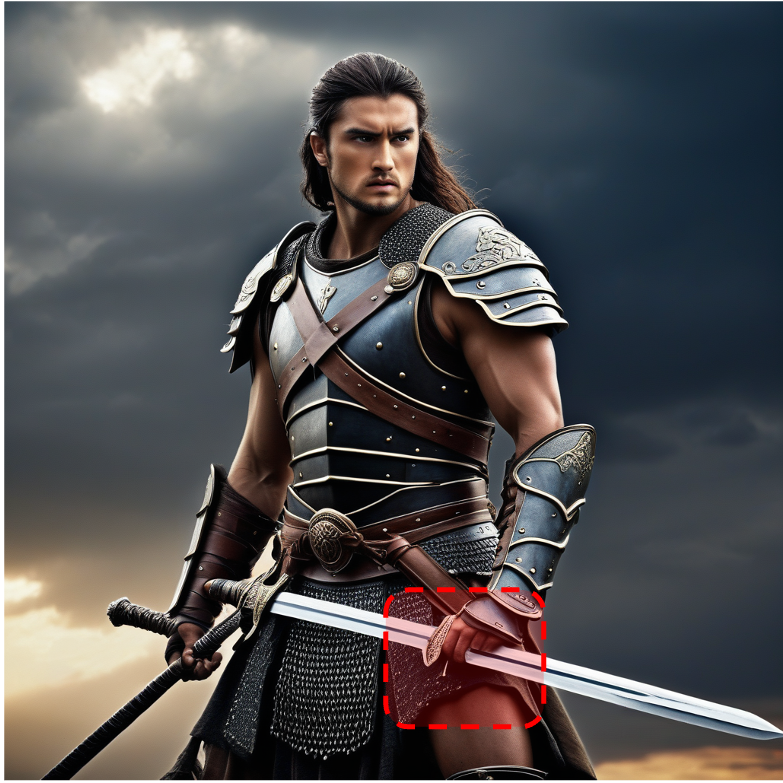}
        \end{subfigure} &
        \begin{subfigure}[b]{0.15\textwidth}
            \centering
            \subcaption{+ Pose-ControlNet}
            \includegraphics[width=\linewidth]{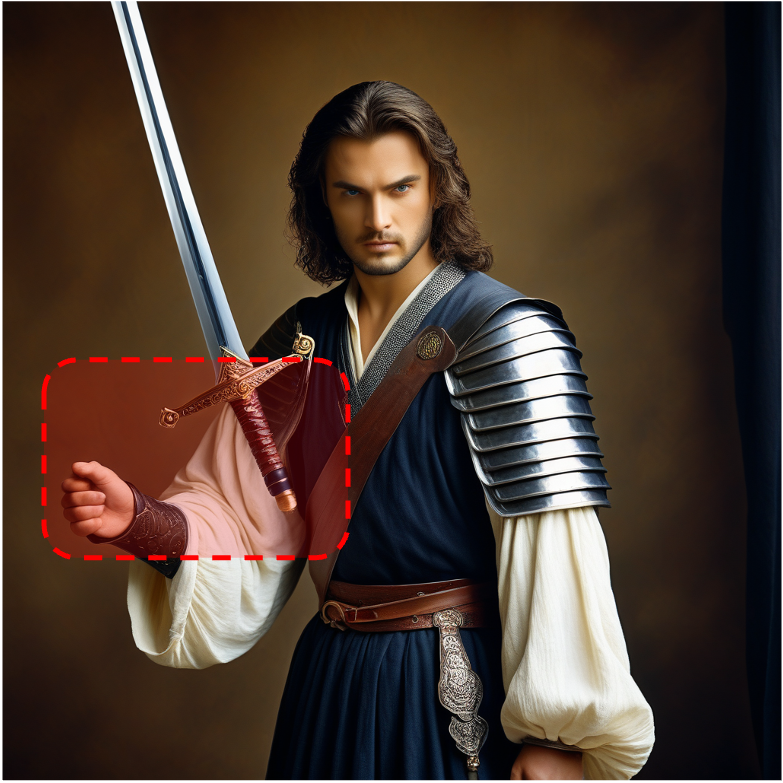}
        \end{subfigure} & 
        \begin{subfigure}[b]{0.15\textwidth}
            \centering
            \subcaption{+ Object Inpainting and HandRefiner}
            \includegraphics[width=\linewidth]{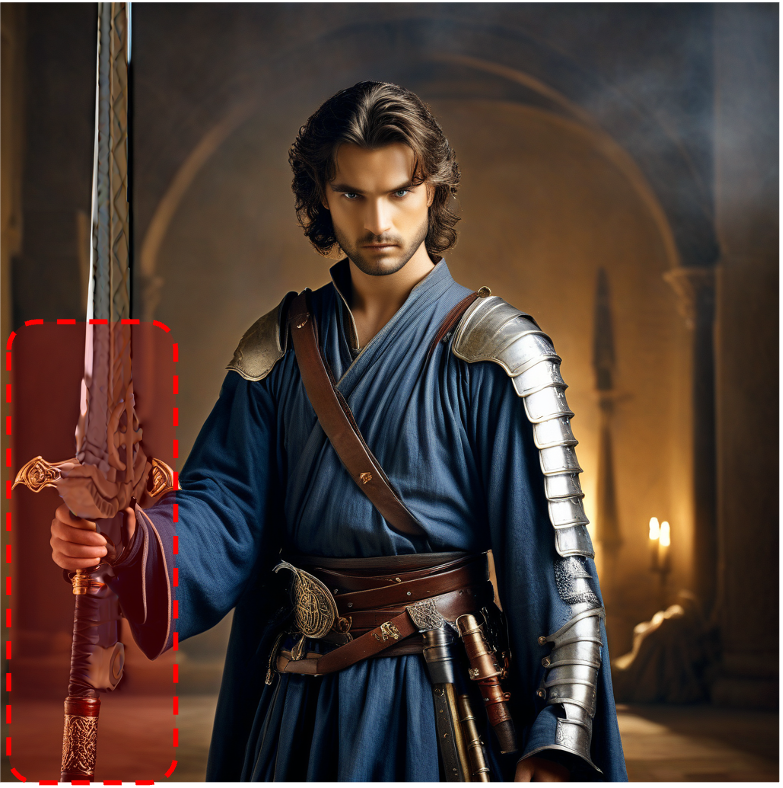}
        \end{subfigure} & 
        \begin{subfigure}[b]{0.15\textwidth}
            \centering
            \subcaption{Ours \\ (w/o Attention Injection)}
            \includegraphics[width=\linewidth]{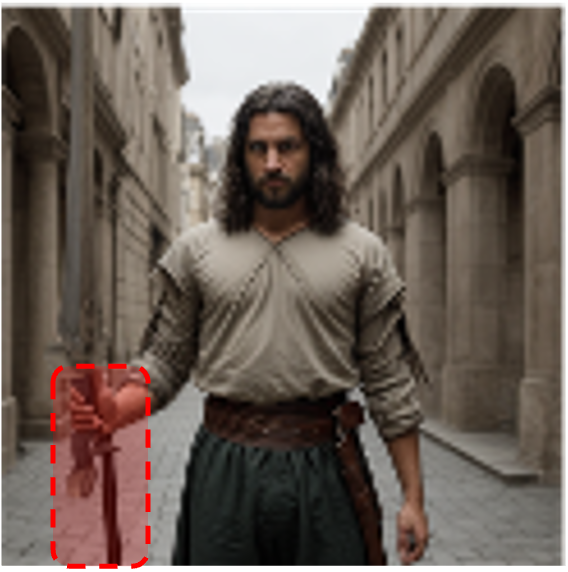}
        \end{subfigure} & 
        \begin{subfigure}[b]{0.15\textwidth}
            \centering
            \subcaption{Ours} 
            \includegraphics[width=\linewidth]{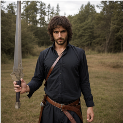}
        \end{subfigure} \\
        \begin{subfigure}[b]{0.15\textwidth}
            \centering
            \includegraphics[width=\linewidth]{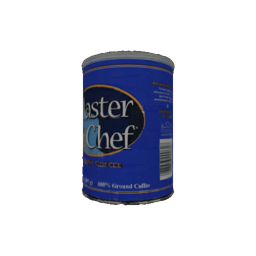}
        \end{subfigure} &
        \begin{subfigure}[b]{0.15\textwidth}
            \centering
            \includegraphics[width=\linewidth]{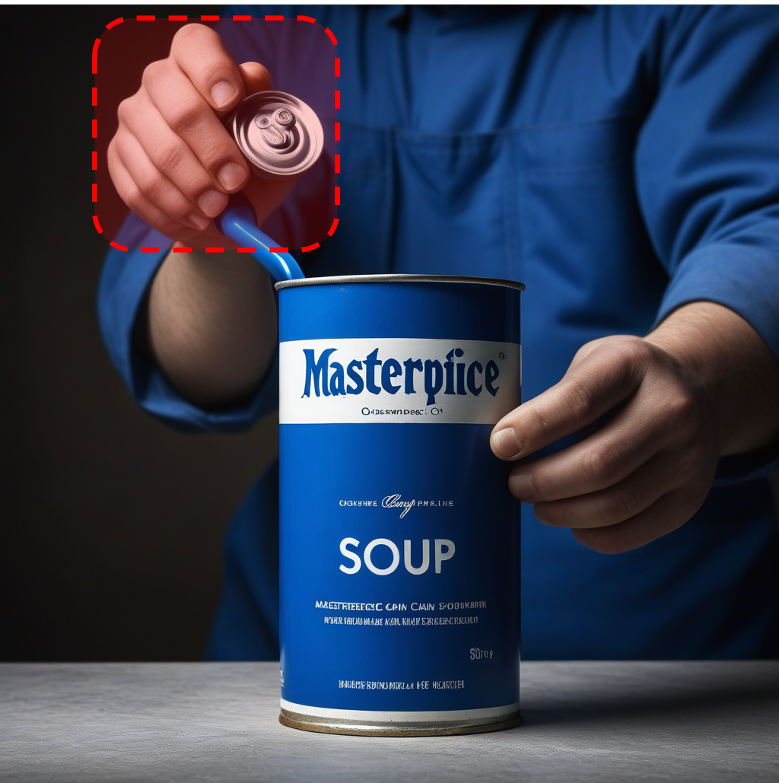}
        \end{subfigure} &
        \begin{subfigure}[b]{0.15\textwidth}
            \centering
            \includegraphics[width=\linewidth]{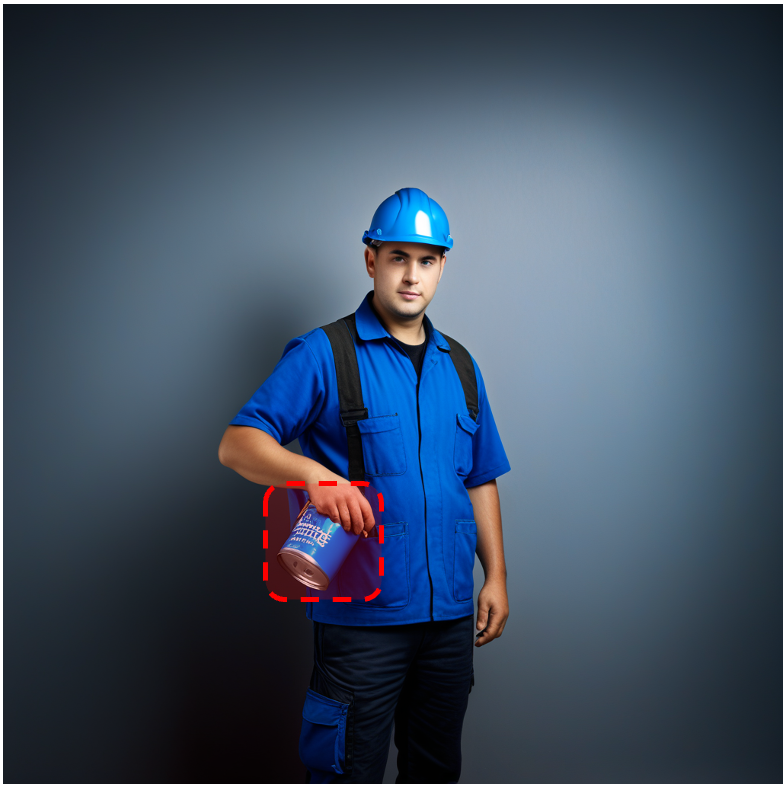}
        \end{subfigure} & 
        \begin{subfigure}[b]{0.15\textwidth}
            \centering
            \includegraphics[width=\linewidth]{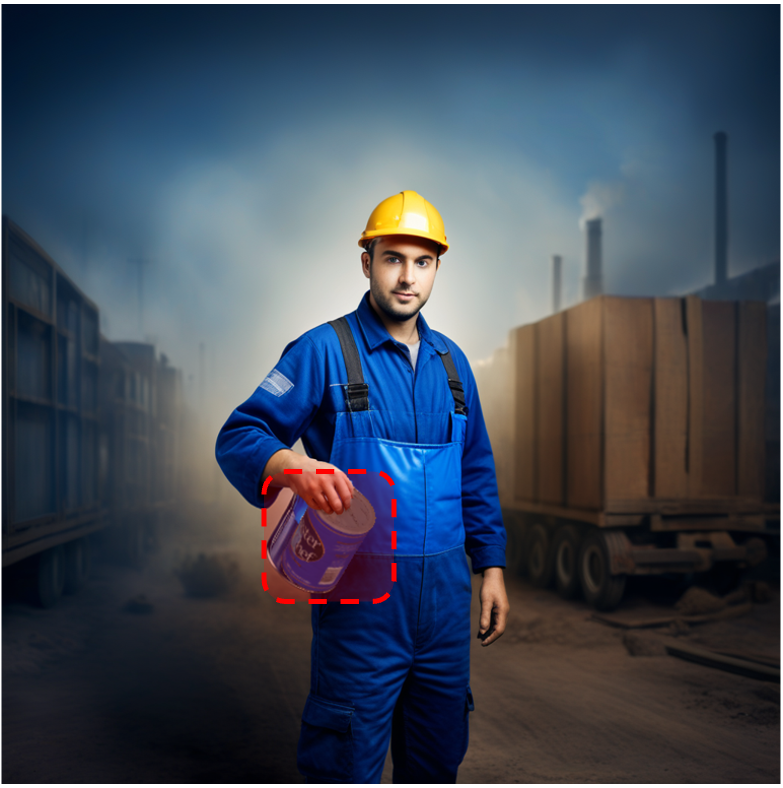}
        \end{subfigure} & 
        \begin{subfigure}[b]{0.15\textwidth}
            \centering
            \includegraphics[width=\linewidth]{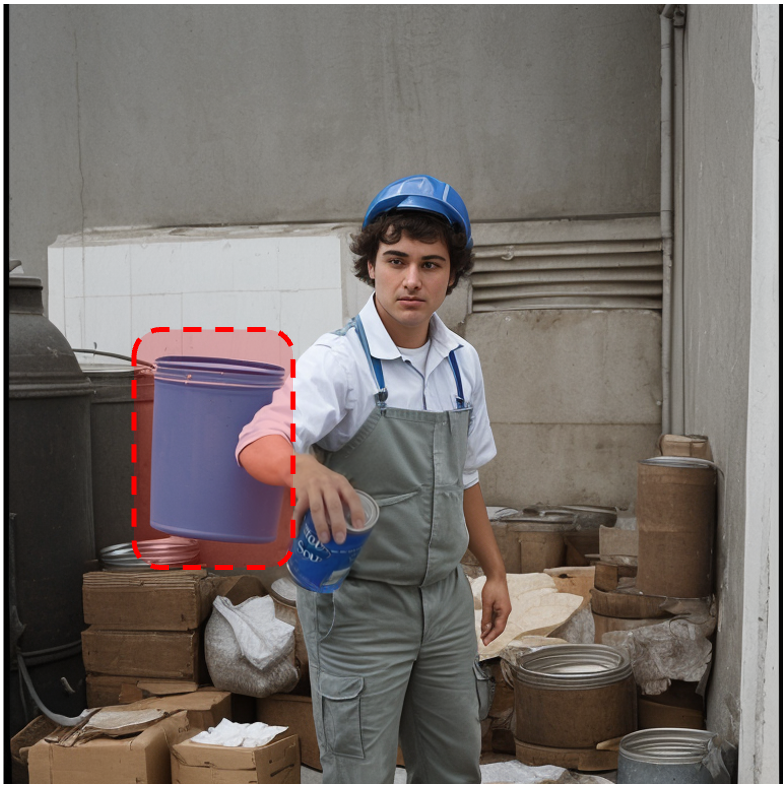}
        \end{subfigure} & 
        \begin{subfigure}[b]{0.15\textwidth}
            \centering
            \includegraphics[width=\linewidth]{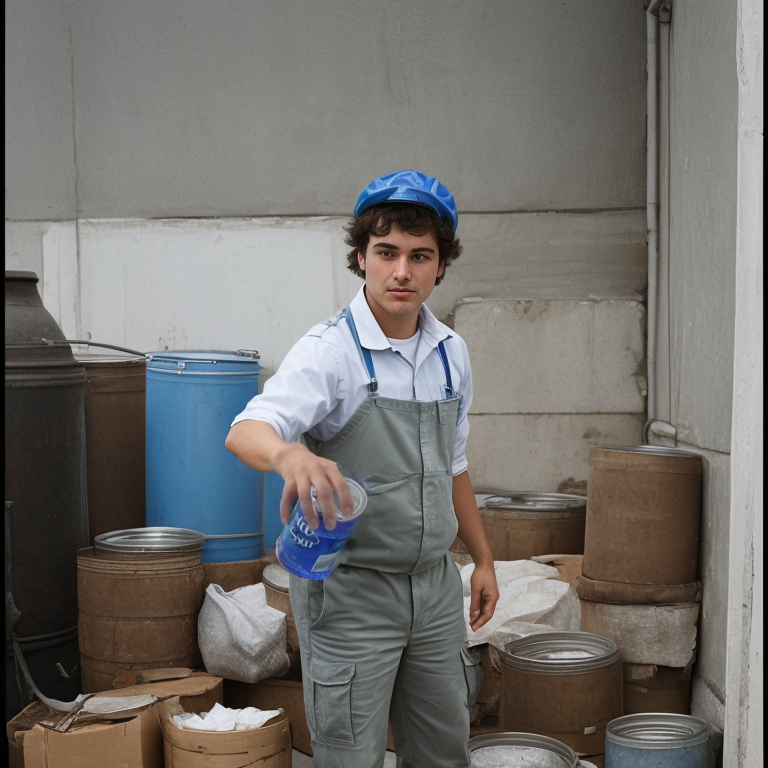}
        \end{subfigure}
   \end{tabular}%
   }
   \vspace{-0.25cm}
  \caption{Qualitative results. We compare HOI images generated by different methods based on a input object (first column). Note that except for the second column, all images were based on the same human pose and object location created from our grasping pipeline. While other methods display erroneous interactions (e.g. multiple objects, object appearance distorted, physically implausible interactions, color blending), which are marked with red segments, our scene-generation pipeline can correctly convey the interaction intention from the full-body grasping pipeline.}
  \label{fig:comp1}
  \vspace{-0.25cm}
\end{figure*}

\subsection{Implementation Details}
\label{sec:formatting}

For the full-body grasping pipeline, we train the body-pose diffusion model on the GRAB \cite{Taheri2020GRABAD} dataset, which captures the full-bodied 3D SMPL-X interaction sequences with various objects. After downsampling the motion sequences of GRAB to 30 fps, we collect all frames that has more than 40 contacting vertices between the object and the subject's right hand, to collect feasible grasps for the training. We then  adopt the Adam optimizer \cite{Kingma2014AdamAM} with a learning rate of $5 \times 10^{-4}$. We train the model with batch size of 2,048 for 50k steps, using 2 RTX 6000 GPUS. For the diffusion schedule, we adapt a cosine noise schedule with $T=1000$. 

\begin{figure}[t]
    \centering
    \resizebox{\columnwidth}{!}{%
    \begin{tabular}{ccc}
        \begin{subfigure}[b]{0.32\columnwidth}
            \centering
            \subcaption{Realistic Vision}
            \includegraphics[width=\textwidth]{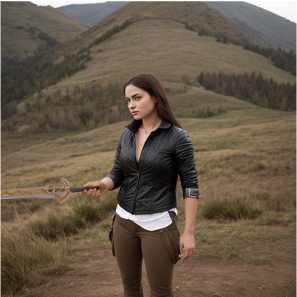}
        \end{subfigure} &
        \hfill
        \begin{subfigure}[b]{0.32\columnwidth}
            \centering
            \subcaption{ToonYou}
            \includegraphics[width=\textwidth]{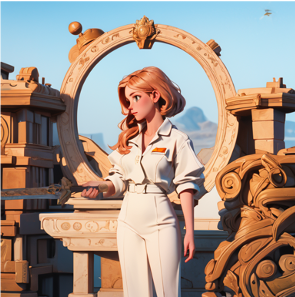}
        \end{subfigure} &
        \hfill
        \begin{subfigure}[b]{0.32\columnwidth}
            \centering
            \subcaption{OrangeMixs}
            \includegraphics[width=\textwidth]{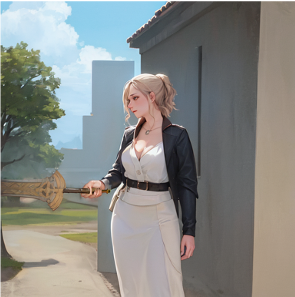}
        \end{subfigure} \\
        \begin{subfigure}[b]{0.32\columnwidth}
            \centering
            \includegraphics[width=\textwidth]{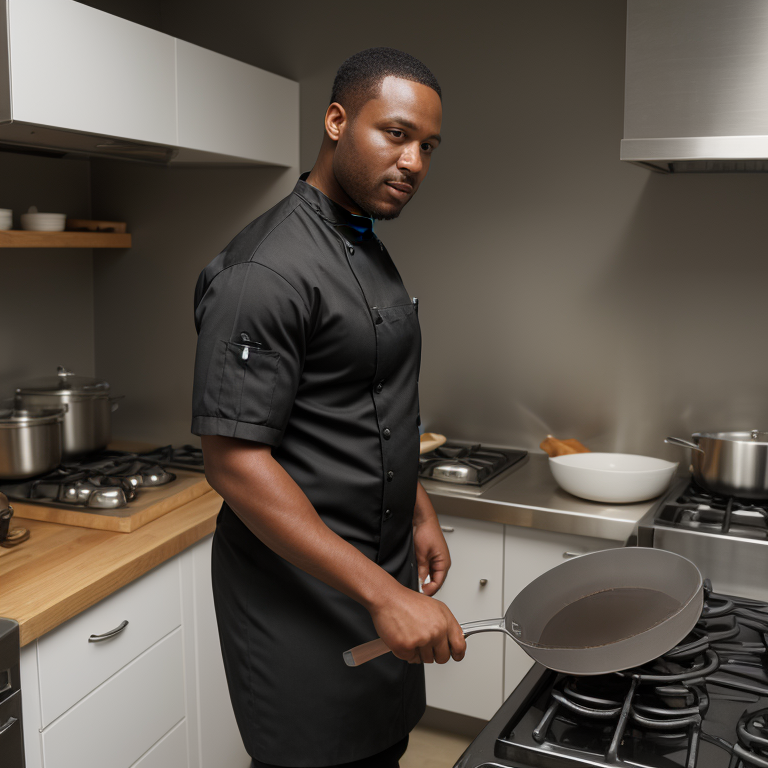}
        \end{subfigure} &
        \hfill
        \begin{subfigure}[b]{0.32\columnwidth}
            \centering
            \includegraphics[width=\textwidth]{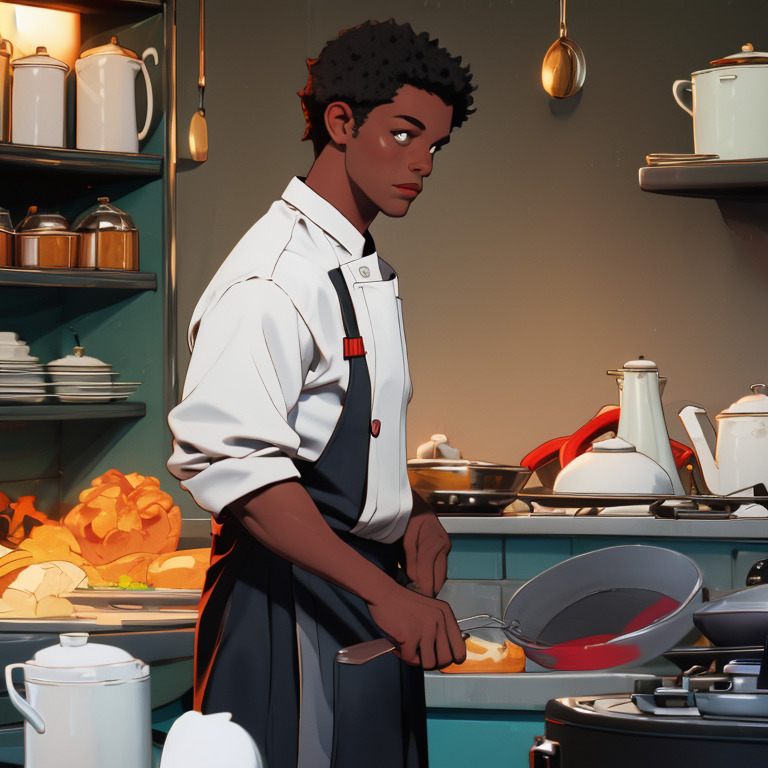}
        \end{subfigure} &
        \hfill
        \begin{subfigure}[b]{0.32\columnwidth}
            \centering
            \includegraphics[width=\textwidth]{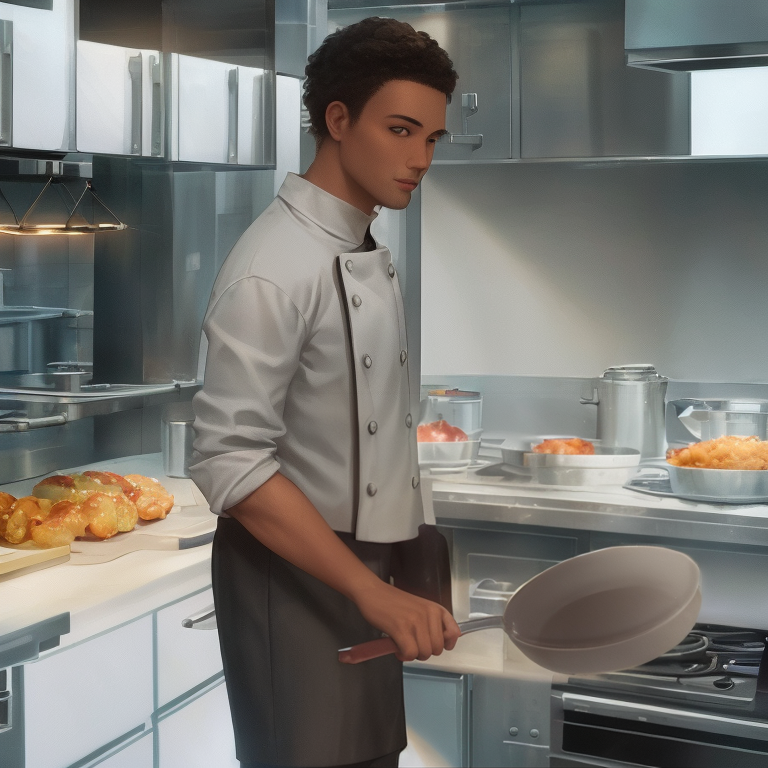}
        \end{subfigure}
    \end{tabular}%
    } 
  \caption{Example results from different models. We display generation results from our pipeline with the same object and body pose, but with different personalized Stable Diffusion models that were acquired from CivitAI \cite{CivitAI} and Huggingface \cite{Huggingface}.}
  \label{fig:diverse_model}
  \vspace{-0.5cm}
\end{figure}

To implement the scene generation pipeline, we train the two modules with the aforementioned custom datasets, using 20K of the interaction pairs. Employing the Stable Diffusion v1.5 \cite{Rombach2021HighResolutionIS} as a base model with parameters frozen, we train the conditional modules with a constant learning rate of $10^{-4}$, for 200 epochs on four A100 GPUS which costs approximately 28 hours. For inference, we used a linear multistep scheduler \cite{Karras2022edm} with 30 inference steps using a classifier-free guidance \cite{ho2021classifierfree} of 3.5. We also support inference using personalized Stable Diffusion models other than the Stable Diffusion v1.5 model used during training, and display results with different models in Fig. \ref{fig:diverse_model}.

\subsection{Quantitative Results}

\textbf{Full-body generation quality} To assess the generation quality, we adopt Frechet Inception Distance (FID) \cite{Heusel2017FID} and Kernel Inception Distance (KID) \cite{Binkowski2018KID}. We compare our results with three baseline models: (1) a finetuned Stable Diffusion v1.5 model \cite{Rombach2021HighResolutionIS} conditioned only by the text description; (2) ControlNet \cite{Zhang_2023_ICCV} with multiple control input; (3) Champ \cite{zhu2024champ}, a human image animation method that uses SMPL-X sequences. For Champ, we separately generate a human image as reference and control the body pose of the reference image using Champ's guidance encoders, without using its motion module. We used the 5K testing set from our novel human-object dataset for comparison. In addition, to assess the alignment between the intended text prompt and the generated image's interaction context, we use CLIPScore \cite{hessel-etal-2021-clipscore} as an additional evaluation metric. As shown in Table \ref{tab:comparisonfid}, our method can improve image quality and prompt alignment in generating images with human-object interaction.

\begin{table}[t]
    \caption{Grasping pose evaluation}
    \label{tab:abalation3}
    \vspace{-0.25cm}
    \centering
    \resizebox{\columnwidth}{!}{%
        \begin{tabular}{lccc}
            \toprule
            Methods & Contact ratio $\uparrow$ & Pose Valid Error $\downarrow$ & Displacement $\downarrow$ \\
            \midrule
            GOAL \cite{Taheri2021GOALG4} & 0.461 & 0.504 & 6.135 \\
            FLEX \cite{Tendulkar2022FLEXFG} & 0.540 & 0.252 & 7.794 \\
            COOP \cite{Zheng_2023_ICCV} & 0.841 & 0.239 & 4.679 \\
            Ours & \textbf{0.909} & \textbf{0.111} & \textbf{2.696} \\
            \bottomrule
        \end{tabular}%
    }
    \vspace{-0.25cm}
\end{table}

\textbf{Hand-grasp generation quality} We also assess the quality of hand-centric images from the hand refinement pipeline, based on both image quality and plausible hand-object pose. To measure instances of successful hand-object contact, we adopt the contact evaluation setup in Affordance Diffusion \cite{Ye2023AffordanceDS} and utilize a widely used hand-object detector \cite{rong2021frankmocap} to measure the object's contact status. We compare our results with three baseline models : (1) a depth-based ControlNet, (2) HandRefiner \cite{lu2023handrefiner}, and (3) Affordance Diffusion \cite{Ye2023AffordanceDS}. We evaluated on a subset of the DexYCB dataset \cite{Chao2021DexYCBAB}. We report the results in Table \ref{tab:comparisonhand}, which shows that our method is capable of outperforming previous methods in creating hand-object images with accurate contact.


\textbf{3D pose evaluation} To evaluate the plausibility of grasping poses for different objects and positions, we constructed a test set of unseen objects distributed far from the original range of the training dataset. We choose 10 novel 3D objects from Dex-YCB \cite{Chao2021DexYCBAB} and 10 human body shapes, and for each pair, we position the object at 64 random 3D positions, relative to the human body's pelvis joint position. Specifically, the x-coordinate (the horizontal position in our paper) ranges from -0.5m to 0.5m, the y-coordinate (the vertical position in our paper) ranges from -0.8m to 0.8m, and the z-coordinate (the direction where the human model is facing) ranges from 0.0m to 0.8m, with the pelvis joint position as its origin. 

For pose evaluation, we utilize VPoser \cite{SMPL-X:2019} to measure the L2 loss of vertex reconstruction from the body poses as a pose-valid error, given that an implausible body pose will result in a higher pose-valid error. We compare our approach with two prior methods that are trained on the GRAB dataset \cite{Taheri2020GRABAD} and support generating full-body grasps for different object translations; GOAL \cite{Taheri2021GOALG4} and COOP \cite{Zheng_2023_ICCV}. For GOAL, we only evaluate the grasping pose generation with optimization (GNet) and set the x-coordinate of the object translation to 0 due to the fact that GOAL does not work when the objects are out of distribution in the horizontal plane \cite{Tendulkar2022FLEXFG}. We present the results in Table ~\ref{tab:abalation3}. The results demonstrate that our model is capable of generating authentic grasping poses for objects in various positions.

\begin{table}[t]
    \caption{Ablation studies on architecture choice}
    \vspace{-0.25cm}
    \centering
    \begin{tabular}{lc}
        \toprule
        Methods & FID $\downarrow$ \\
        \midrule
        Ours & \textbf{22.88} \\
        \midrule
        w/o object rendering & 29.53 \\
        w/o human skeleton & 26.35 \\
        w/o joint depth & 24.37  \\
        \bottomrule
    \end{tabular}
    \label{tab:abalation1}
  \vspace{-0.5cm}
\end{table}

\textbf{Ablation Studies} During the scene generation pipeline, we utilize different structural renderings from the 3D grasping model as conditions to generate a realistic image; the object rendering defines the appearance and location of the target object, the skeleton map gives a precise human pose depiction, and the depth map maintains geometric consistency. To investigate the importance of each factor, we measure the FID scores for cases where only two of the three conditional modules are provided during inference. As presented in Table ~\ref{tab:abalation1}, our full model setting outperforms other settings with missing conditions. 

In Table~\ref{tab:comparisonfid}, we also display results for the main setting without attention injection, which alleviates the risk of generating erroneous interactions. While the setting without attention injection has a slightly better FID / KID score, our full setting shows a substantial increase in CLIPScore, signifying that the generated images successfully adhere to the given interaction context.

\textbf{User Study} We conducted a user study to measure the perceptual quality and geometric consistency of our pipeline. We asked 28 participants to compare images that were generated based on the same 3D grasping pose and object, one generated using multiple ControlNets \cite{Zhang_2023_ICCV} and HandRefiner \cite{lu2023handrefiner} and the other using GraspDiffusion. The participants were asked two types of questions: (a) which image is more realistic and plausible, and (b) given the original grasping information, which one follows faithfully to the grasping context. In total, 92.4\% of the votes preferred our method over the baseline on plausibility, while 96.4\% preferred based on following the given context.

\subsection{Applications}

By utilizing a 3D full-body grasping model generated from an object input, GraspDiffusion can provide a practical solution for AI practitioners who intend to use generative AI for their artwork such as advertisements, illustrations, and comic books. To alleviate the requirement of an object mesh model, our pipeline also supports using 3D reconstruction models \cite{TripoSR2024, wu2024unique3d, Long2023Wonder3DSI} that can recover 3D mesh models from a single image input. Moreover, our pipeline can support various personalized image domains, including (but not limited to) realistic, anime, pixel art style, and more. In Figure~\ref{fig:diverse_model} we present results from the same 3D grasping pose, using diverse personalized text-to-image models to support different art styles and backgrounds. Further examples are provided in the Supplementary Material.

%% file: sec/5_conclusion.tex
\section{Discussions and Limitations}

While GraspDiffusion can produce humans with detailed finger articulation and accurate object interaction, several samples exhibit discrepancies between the body's skin texture and the refined hand's skin texture. We account this issue to the shortage of balanced, high-quality data samples during training, and opt to construct additional interaction samples to facilitate high-quality generation.
In the future, we aspire to extend our pipeline toward scene-level generation that involves interaction between multiple humans and objects, and to better define interactions in the form of user-controllable text prompts. We also look forward to synthesize zero-shot interaction motions by leveraging image-to-video diffusion models \cite{Blattmann2023StableVD, yang2024cogvideox, HaCohen2024LTXVideo, kong2024hunyuanvideo}.

\section{Conclusion}

We present an image generation pipeline that is the first to explicitly target realistic human-object interaction. The resulting images exhibit both explicit (hand-object contact, realistic hand grasp) and implicit human-object interaction (human gaze, body direction), without requiring any auxiliary conditions other than a 3D object mesh and its relative position. The results demonstrate our method's effectiveness in creating images with plausible hand poses, while preserving the given object's identity. In the future, we plan to extend our pipeline towards generating various types of interaction (e.g. human-human interaction, specialized hand-object interaction), while further demonstrating the effectiveness of our pipeline in video generation and synthetic dataset creation for interaction detection.

%% file: sec/x_supplementary.tex
\section{Model Architecture}
\label{sec:supparch}

\begin{figure*}
    \begin{tabular}{cc}
        \begin{minipage}[b]{0.49\textwidth}
            \centering
            \caption*{Object: an axe}
            \vspace{-0.4cm}
            \includegraphics[width=\textwidth]{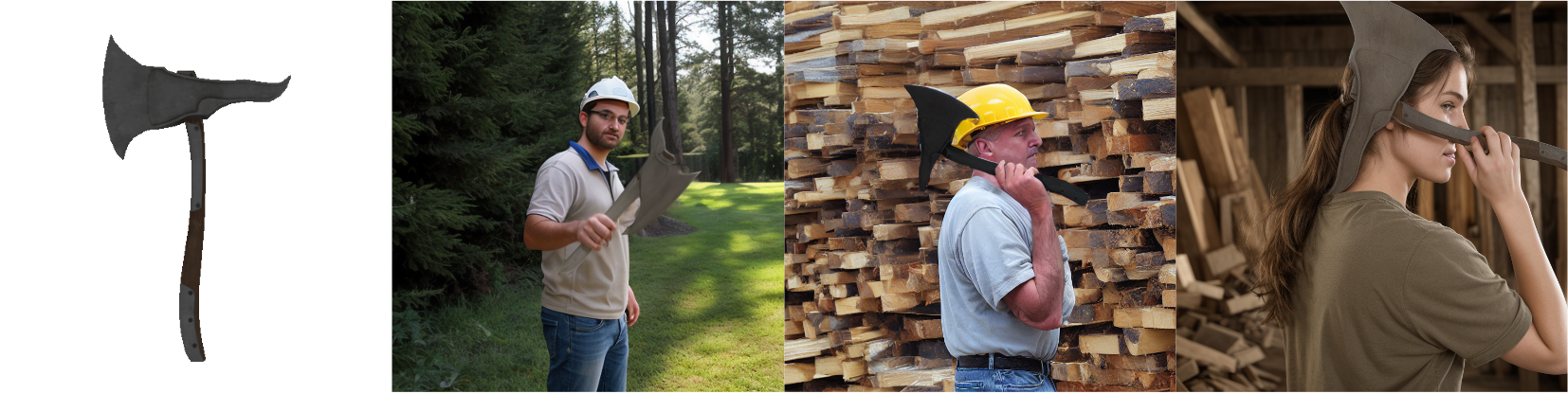}
        \end{minipage} &
        \begin{minipage}[b]{0.49\textwidth}
            \centering
            \caption*{Object: a red mug cup}
            \vspace{-0.4cm}
            \includegraphics[width=\textwidth]{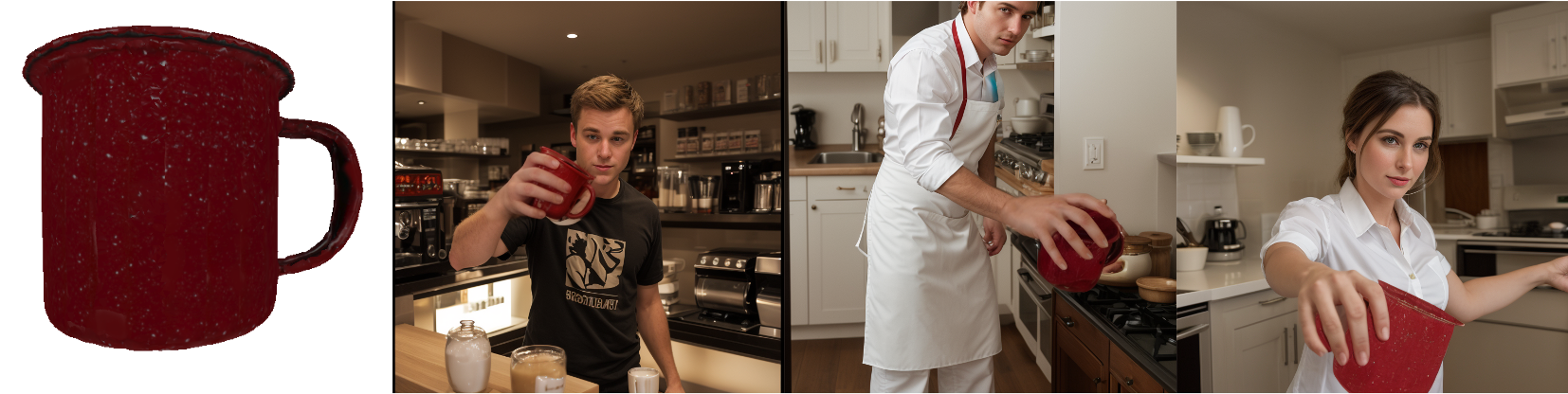}
        \end{minipage} \\
        \begin{minipage}[b]{0.49\textwidth}
            \centering
            \caption*{Object: a red plastic bowl}
            \vspace{-0.4cm}
            \includegraphics[width=\textwidth]{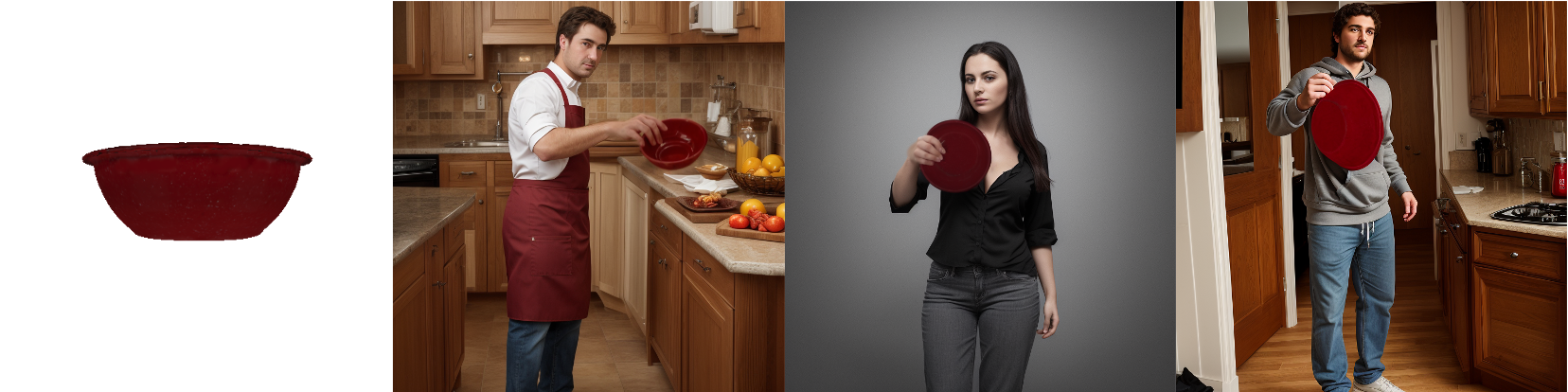}
        \end{minipage} &
        \begin{minipage}[b]{0.49\textwidth}
            \centering
            \caption*{Object: a black wineglass}
            \vspace{-0.4cm}
            \includegraphics[width=\textwidth]{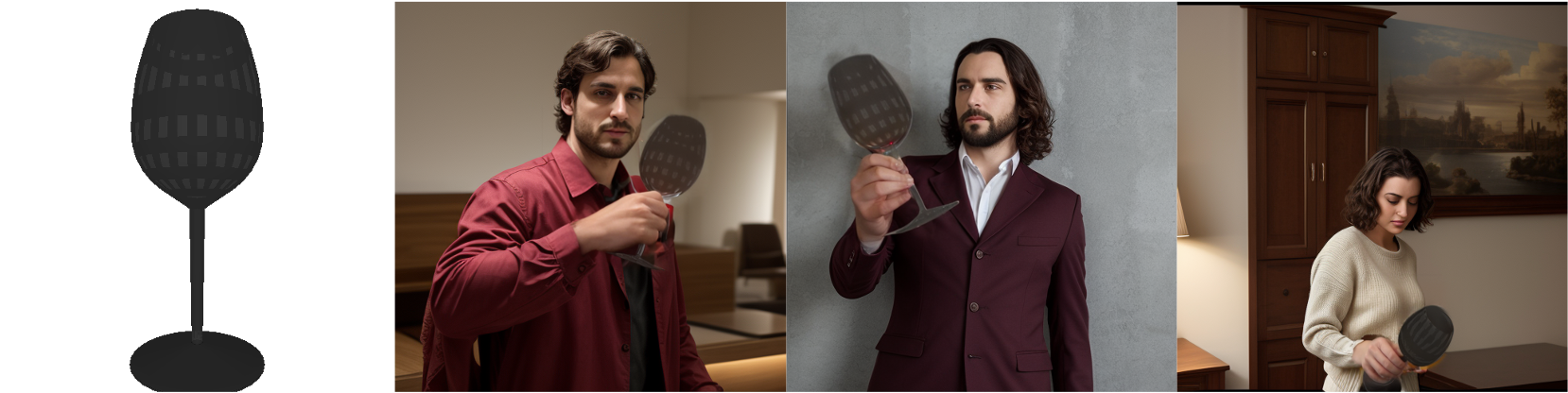}
        \end{minipage} \\
        \begin{minipage}[b]{0.49\textwidth}
            \centering
            \caption*{Object: a black sword}
            \vspace{-0.4cm}
            \includegraphics[width=\textwidth]{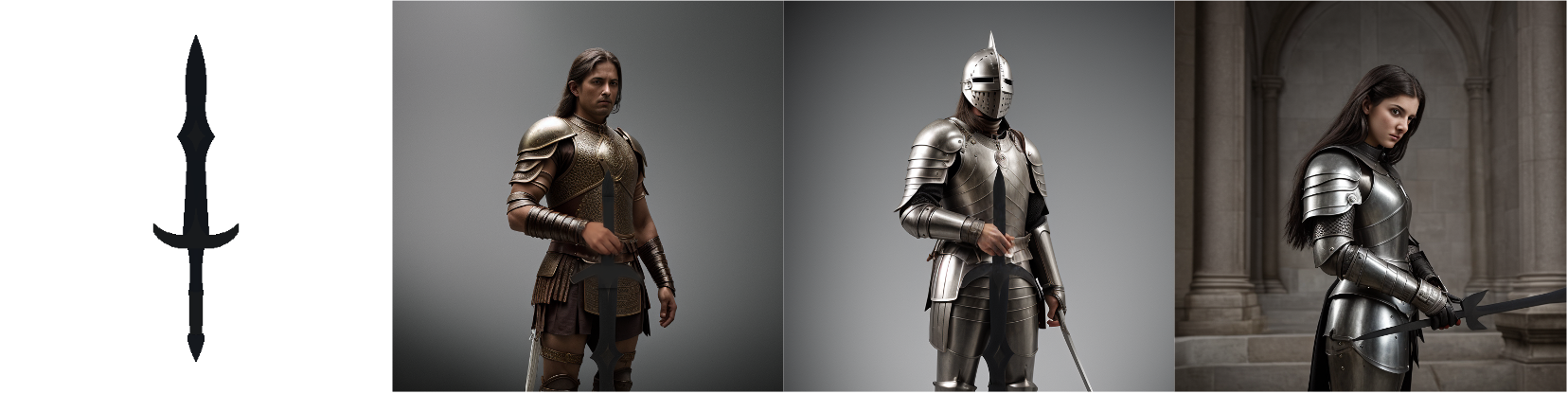}
        \end{minipage} &
        \begin{minipage}[b]{0.49\textwidth}
            \centering
            \caption*{Object: a coffee cup}
            \vspace{-0.4cm}
            \includegraphics[width=\textwidth]{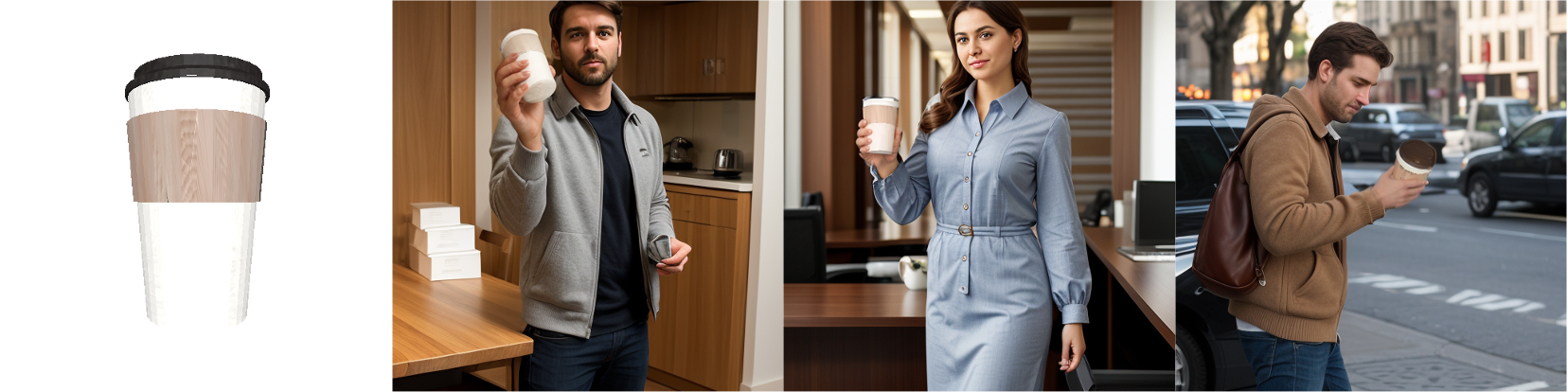}
        \end{minipage} \\
        \begin{minipage}[b]{0.49\textwidth}
            \centering
            \caption*{Object: a dagger}
            \vspace{-0.4cm}
            \includegraphics[width=\textwidth]{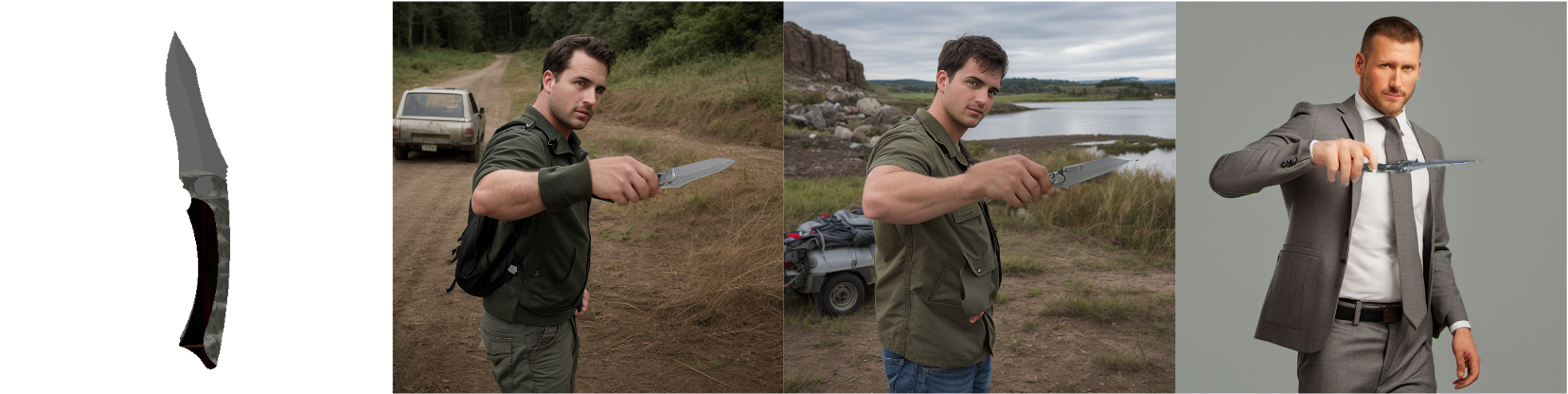}
        \end{minipage} &
        \begin{minipage}[b]{0.49\textwidth}
            \centering
            \caption*{Object: a tomato soup can}
            \vspace{-0.4cm}
            \includegraphics[width=\textwidth]{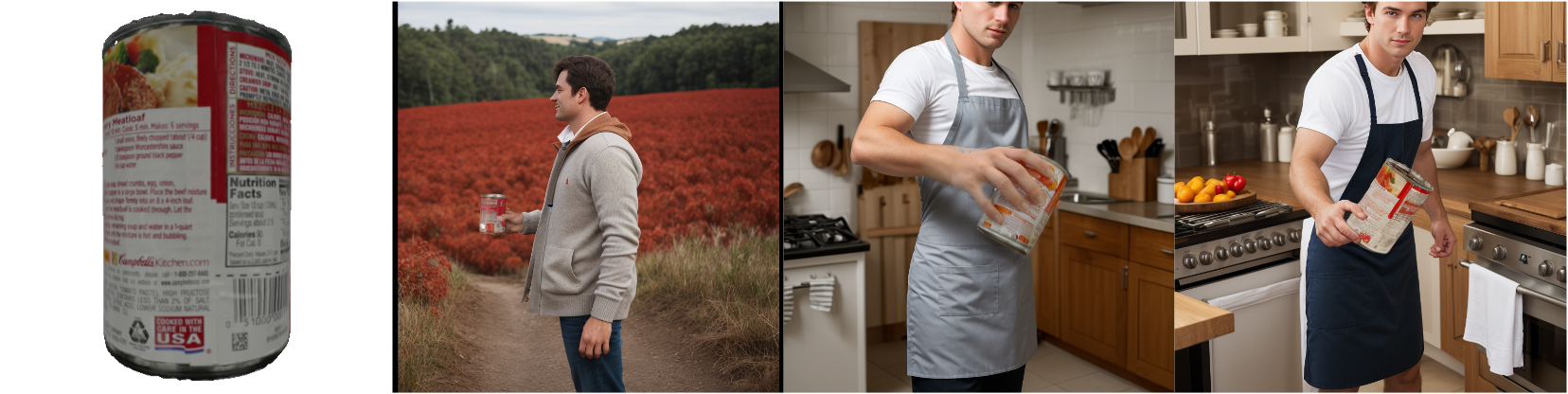}
        \end{minipage} \\
        \begin{minipage}[b]{0.49\textwidth}
            \centering
            \caption*{Object: a mustard bottle}
            \vspace{-0.4cm}
            \includegraphics[width=\textwidth]{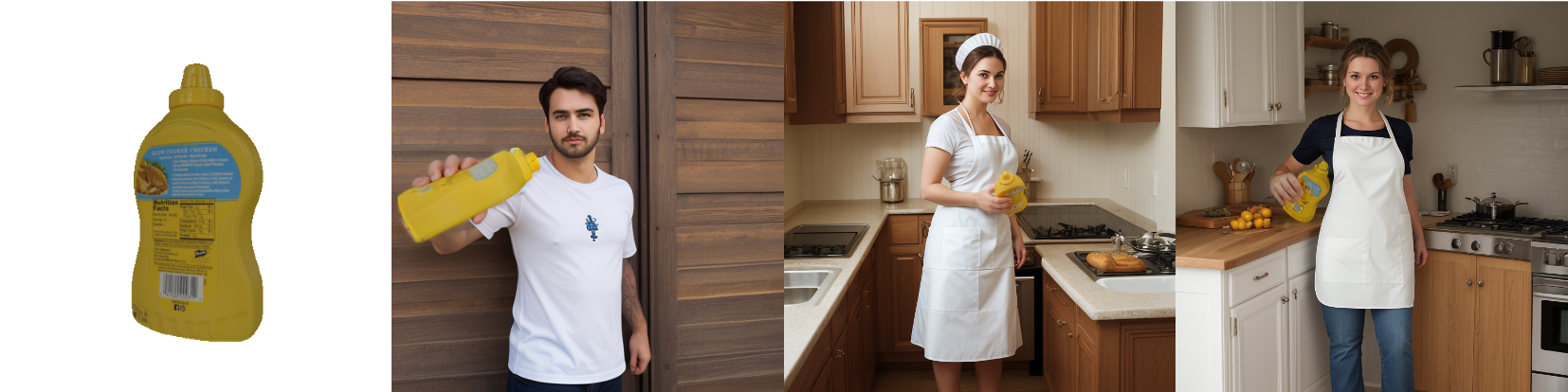}
        \end{minipage} &
        \begin{minipage}[b]{0.49\textwidth}
            \centering
            \caption*{Object: a baseball bat}
            \vspace{-0.4cm}
            \includegraphics[width=\textwidth]{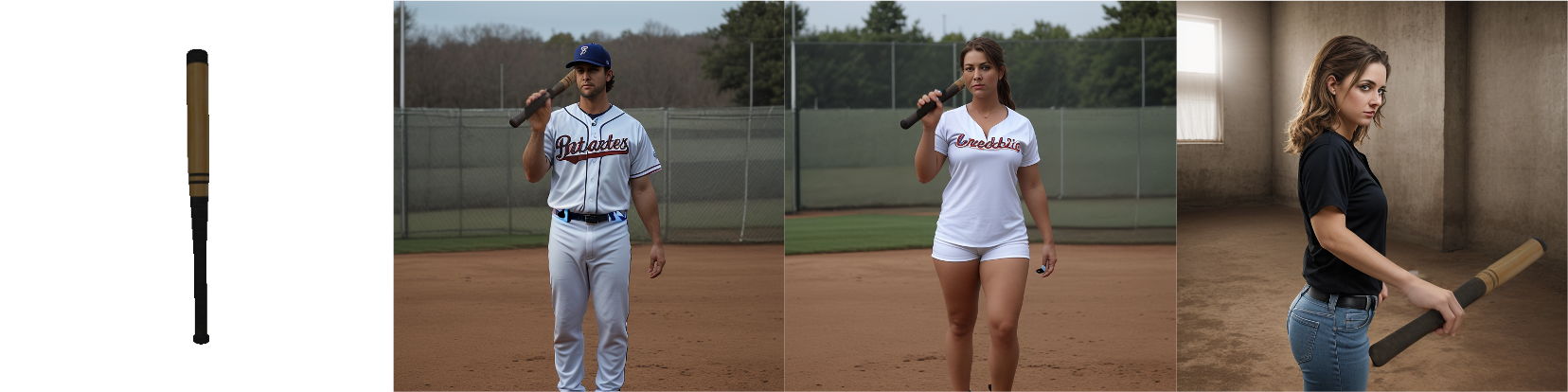}
        \end{minipage} \\
        \begin{minipage}[b]{0.49\textwidth}
            \centering
            \caption*{Object: a can of spam}
            \vspace{-0.4cm}
            \includegraphics[width=\textwidth]{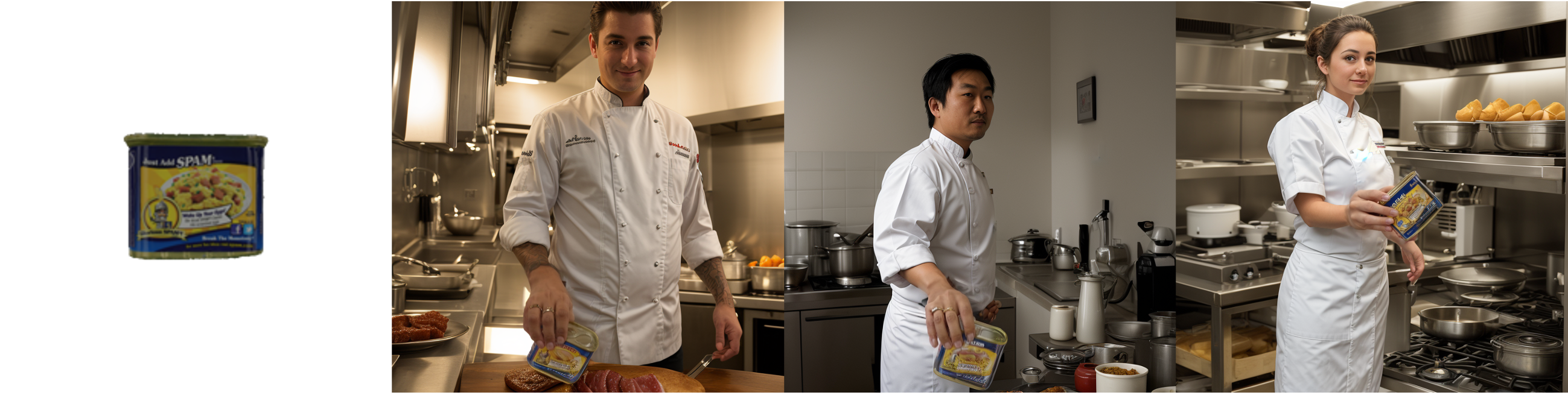}
        \end{minipage} &
        \begin{minipage}[b]{0.49\textwidth}
            \centering
            \caption*{Object: a banana}
            \vspace{-0.4cm}
            \includegraphics[width=\textwidth]{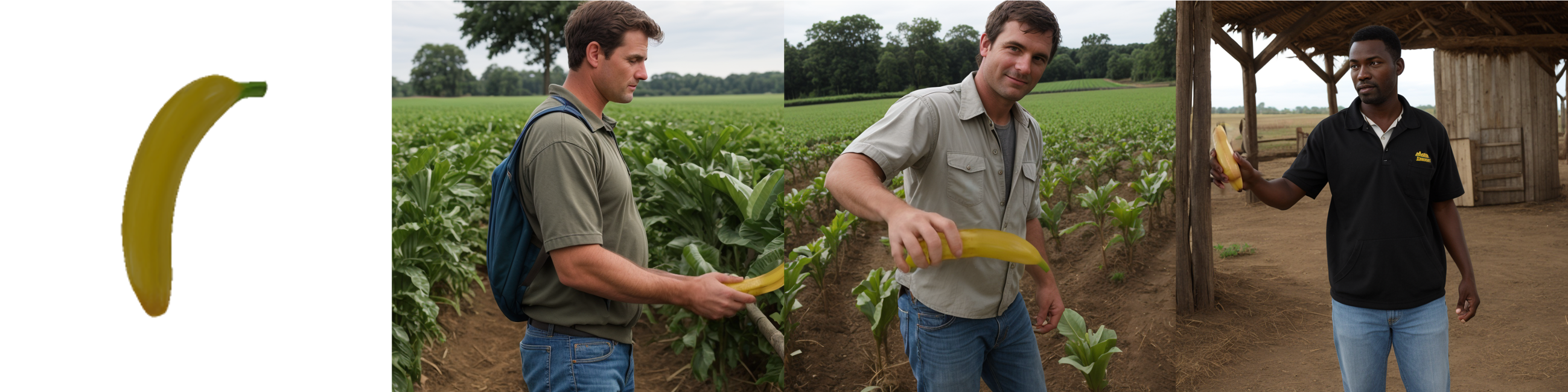}
        \end{minipage} \\
  \end{tabular}
    \captionof{figure}{More results on synthesized images from a given object. Objects were gathered from the DexYCB dataset \cite{Chao2021DexYCBAB} and SketchFab \cite{SketchFab}}
    \label{fig:ab2}
\end{figure*}

For the first stage of our pipeline, we use a diffusion model to synthesize a body pose grasping the input object. The model is trained to predict plausible body parameters (6 DoF body pose, global orientation), conditioned on the object's relative location $t_\text{obj} \in \mathbb{R}^3$ and the target hand $c_\text{left}, c_\text{right} \in \{0,1\} $. These conditions are transformed to a conditional embedding $v_c$, which is added to the timestep embedding $e_t$ and passed to residual blocks within the model, following \cite{pmlr-v139-nichol21a}. We used 3 ResNet blocks for the model, and adopted a cosine noise schedule in training. Additional grasping results are provided in Figure~\ref{fig:ab2}, and details on model parameters are provided in Table~\ref{tab:model1params}.

\begin{table}[b]
    \small
    \centering
    \resizebox{0.8\columnwidth}{!}{%
    \begin{tabular}{lc}
        \toprule
        Parameter & Diffusion Model (Body Pose)  \\
        \midrule
        Input Channels & 132 \\
        Condition Channels & 5 \\
        Model Channels & 1024 \\
        ResBlock Number & 3 \\
        Diffusion Steps & 1000 \\
        Noise Scheduler & Cosine \\
        \bottomrule
    \end{tabular}
    }
    \caption{Model architecture for body pose generation diffusion model.}
    \vspace{-12pt}
    \label{tab:model1params}
\end{table}

\begin{figure}[b]
    \centering
    \resizebox{\columnwidth}{!}{%
    \begin{tabular}{cccc}
        \begin{minipage}[b]{0.24\columnwidth}
            \centering
            \includegraphics[width=\textwidth]{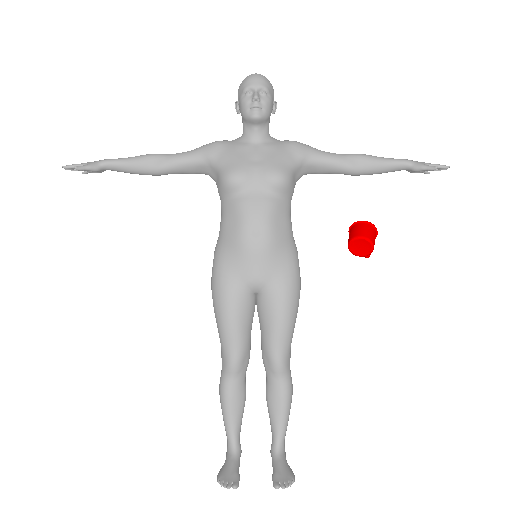}
            \subcaption{Location}
        \end{minipage} &
        \begin{minipage}[b]{0.24\columnwidth}
            \centering
            \includegraphics[width=\textwidth]{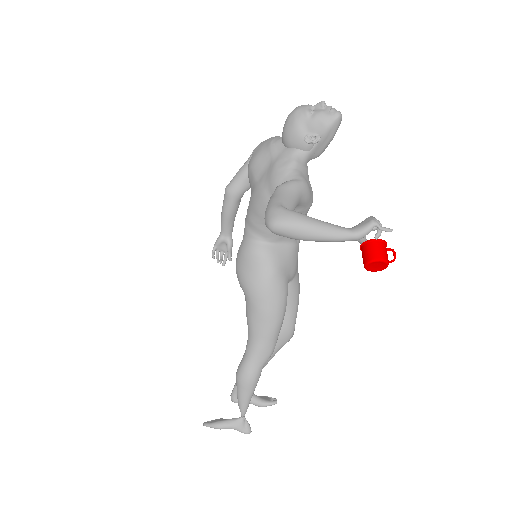}
            \subcaption{GOAL \cite{Taheri2021GOALG4}}
        \end{minipage} &
        \begin{minipage}[b]{0.24\columnwidth}
            \centering
            \includegraphics[width=\textwidth]{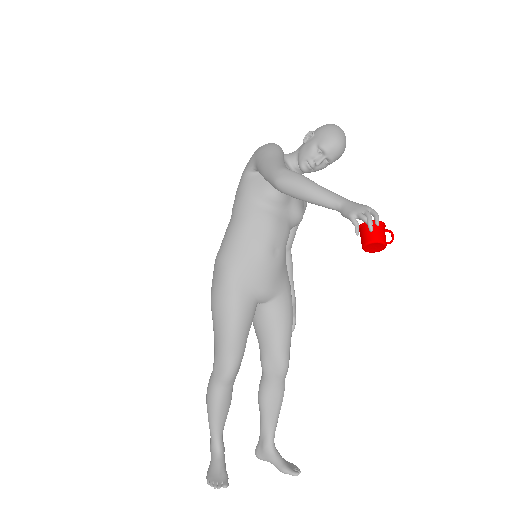}
            \subcaption{COOP \cite{Zheng_2023_ICCV}}        
        \end{minipage} & 
        \begin{minipage}[b]{0.24\columnwidth}
            \centering
            \includegraphics[width=\textwidth]{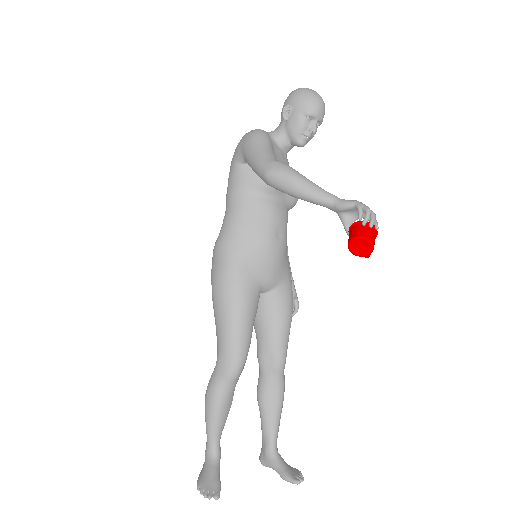}
            \subcaption{Ours}
        \end{minipage} \\
    \end{tabular}%
  }
    \vspace{-10pt}
    \caption{Grasp synthesis comparison with previous methods.}
    \label{fig:rebuttal1}
    \vspace{-10pt}
\end{figure}

In Figure~\ref{fig:rebuttal1}, we provide an example comparison between our method and previous grasping pose generation methods ~\cite{Taheri2021GOALG4, Wu2021SAGASW, Tendulkar2022FLEXFG, Zheng_2023_ICCV}. When given a object with its location relative to the human body (left row), GOAL \cite{Taheri2021GOALG4} and SAGA \cite{Wu2021SAGASW} tend to create distorted poses when the object is far away from the human, as they assume it to be in the same horizontal xy-plane. FLEX assumes a world-centric coordinate system, which leads to pose ambiguity for our scenario. While COOP has a similar objective, it focuses on various object heights, and requires an extensive test-time optimization of 5 different loss terms. We are the first to utilize a lightweight diffusion model in synthesizing body grasping poses.

\begin{figure}[t]
    \centering
    \resizebox{\columnwidth}{!}{%
    \begin{tabular}{ccc}
        \begin{minipage}[b]{0.14\textwidth}
            \centering
            \subcaption{Object Image}
            \includegraphics[width=\textwidth]{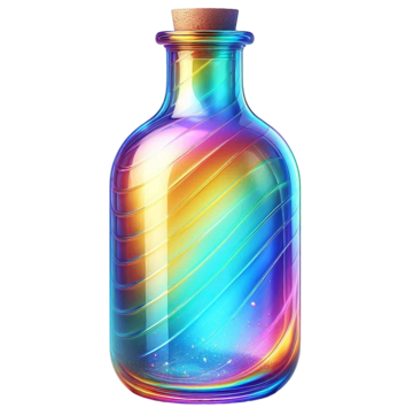}
        \end{minipage} &
        \begin{minipage}[b]{0.14\textwidth}
            \centering
            \subcaption{3D Mesh}
            \includegraphics[width=\textwidth]{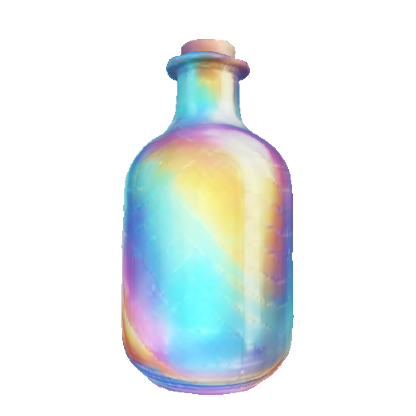}
        \end{minipage} &
        \begin{minipage}[b]{0.14\textwidth}
            \centering
            \subcaption{Result image}
            \includegraphics[width=\textwidth]{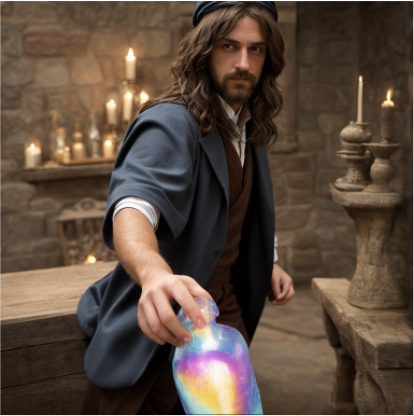}
        \end{minipage} \\
        \begin{minipage}[b]{0.14\textwidth}
            \centering
            \includegraphics[width=\textwidth]{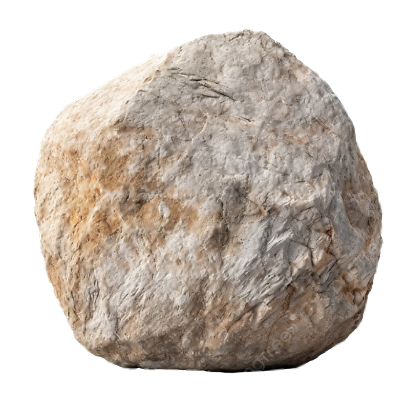}
        \end{minipage} &
        \begin{minipage}[b]{0.14\textwidth}
            \centering
            \includegraphics[width=\textwidth]{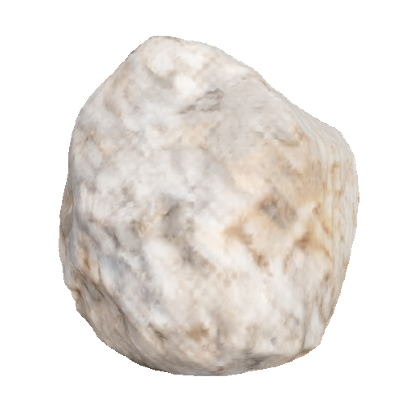}
        \end{minipage} &
        \begin{minipage}[b]{0.14\textwidth}
            \centering
            \includegraphics[width=\textwidth]{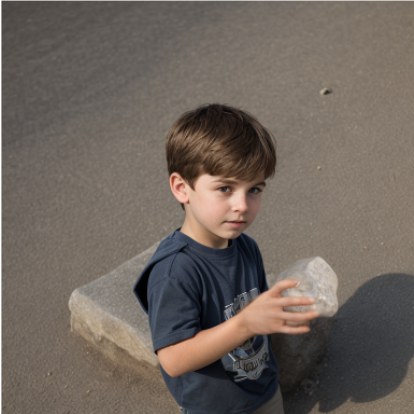}
        \end{minipage} \\
        \begin{minipage}[b]{0.14\textwidth}
            \centering
            \includegraphics[width=\textwidth]{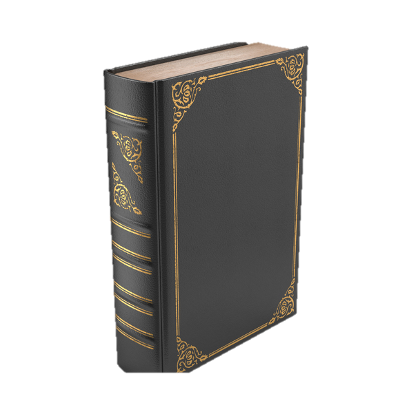}
        \end{minipage} &
        \begin{minipage}[b]{0.14\textwidth}
            \centering
            \includegraphics[width=\textwidth]{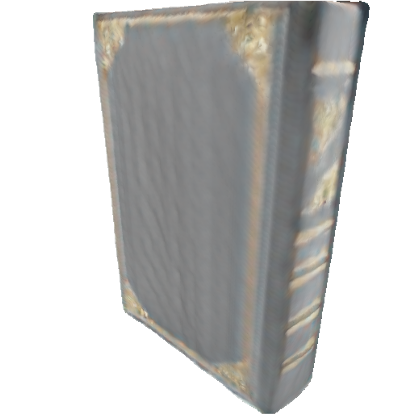}
        \end{minipage} &
        \begin{minipage}[b]{0.14\textwidth}
            \centering
            \includegraphics[width=\textwidth]{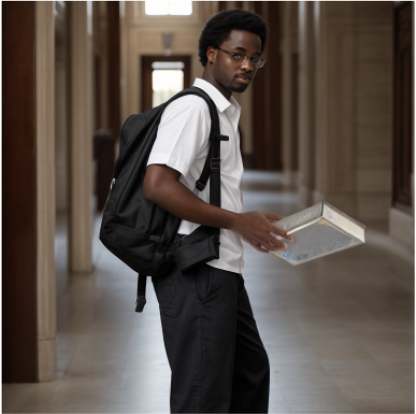}
        \end{minipage} \\
        \begin{minipage}[b]{0.14\textwidth}
            \centering
            \includegraphics[width=\textwidth]{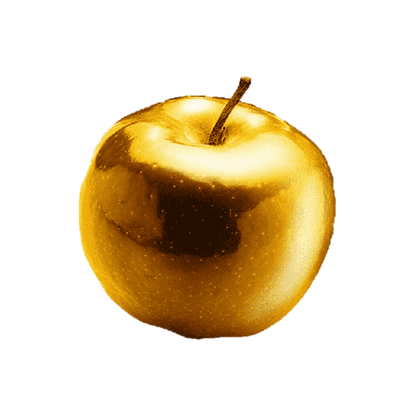}
        \end{minipage} &
        \begin{minipage}[b]{0.14\textwidth}
            \centering
            \includegraphics[width=\textwidth]{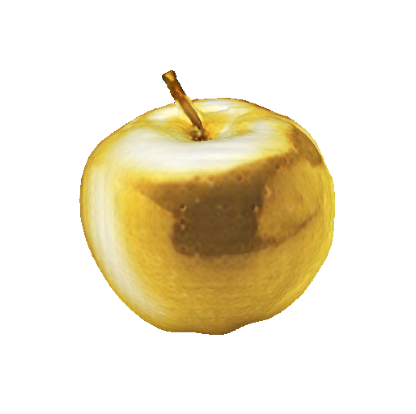}
        \end{minipage} &
        \begin{minipage}[b]{0.14\textwidth}
            \centering
            \includegraphics[width=\textwidth]{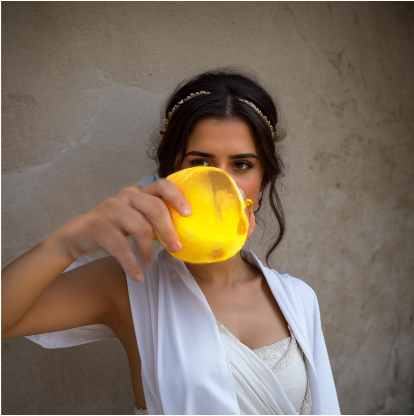}
        \end{minipage} \\
    \end{tabular}%
  }
  \caption{Synthesis results from a single image. We first synthesize a 3D Mesh from the image using TripoSR \cite{TripoSR2024}, InstantMesh \cite{xu2024instantmesh}, Real3D \cite{jiang2024real3d} and subsequently used the mesh as input.}
  \label{fig:img23d_sample}
  \vspace{-0.5cm}
\end{figure}
  
\begin{table}[b]
    \small
    \centering
    \resizebox{0.8\columnwidth}{!}{%
    \begin{tabular}{lc}
        \toprule
        Parameter & Conditional Encoder(s) \\
        \midrule
        Input Channels & $3 \times 64$ \\
        Output Channels & $[320, 640, 1280, 1280]$ \\
        ResBlock Number & 2 \\
        Kernel Size & 1 \\
        Feature Weight (Body) & $[1.0, 0.6, 1.0]$ \\
        Feature Weight (Hand) & $[1.0, 0.6, 1.0]$ \\
        \midrule
        Parameter & Attention Injection   \\
        \midrule
        Human Strength & 0.2 \\
        Object Strength & 1.8 \\
        Negative Object Strength & -9.0 \\
        Weight coefficient $(w^{'})$ & 0.4 \\
        \bottomrule
    \end{tabular}
    }
    \caption{Model architecture for scene generation models, and inference parameters for attention injection. 
    }
    \vspace{-12pt}
    \label{tab:model2params}
\end{table}

\begin{figure}[t]
    \centering
    \resizebox{\columnwidth}{!}{%
    \begin{tabular}{ccc}
    \begin{minipage}[b]{0.32\columnwidth}
        \centering
        \includegraphics[width=\textwidth]{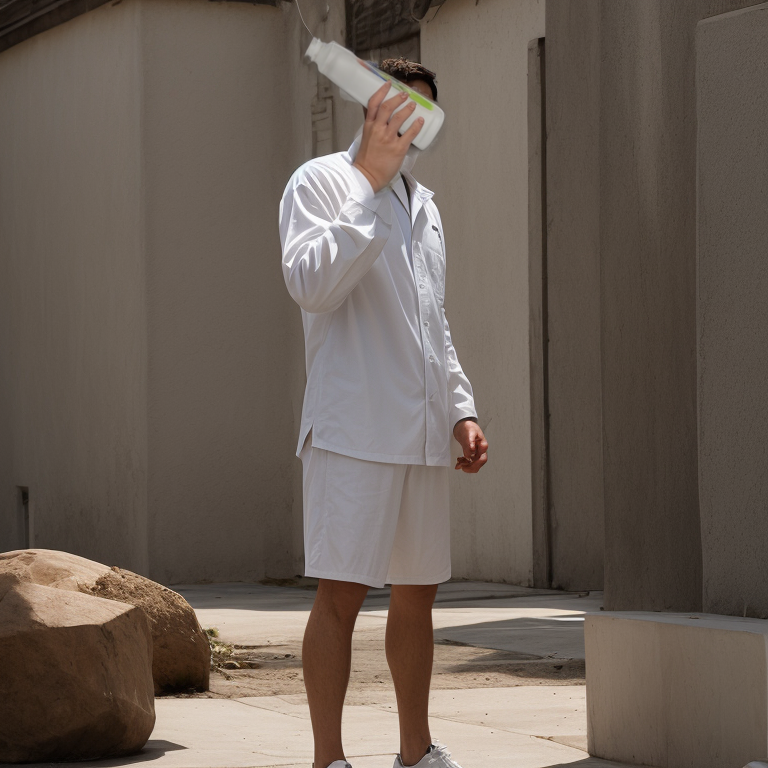}
    \end{minipage} &
    \begin{minipage}[b]{0.32\columnwidth}
        \centering
        \includegraphics[width=\textwidth]{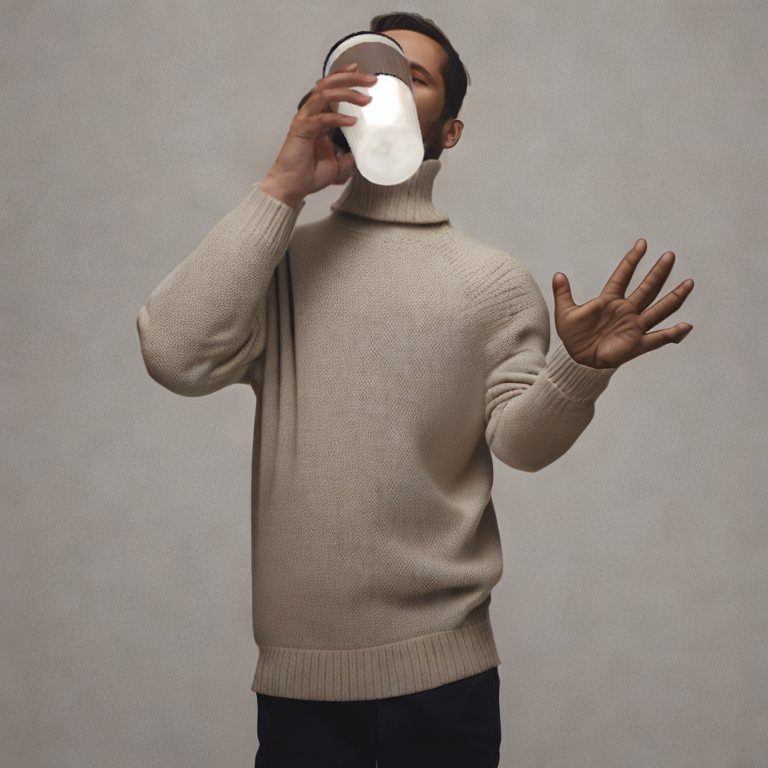}
    \end{minipage} &
    \hfill
    \begin{minipage}[b]{0.32\columnwidth}
        \centering
        \includegraphics[width=\textwidth]{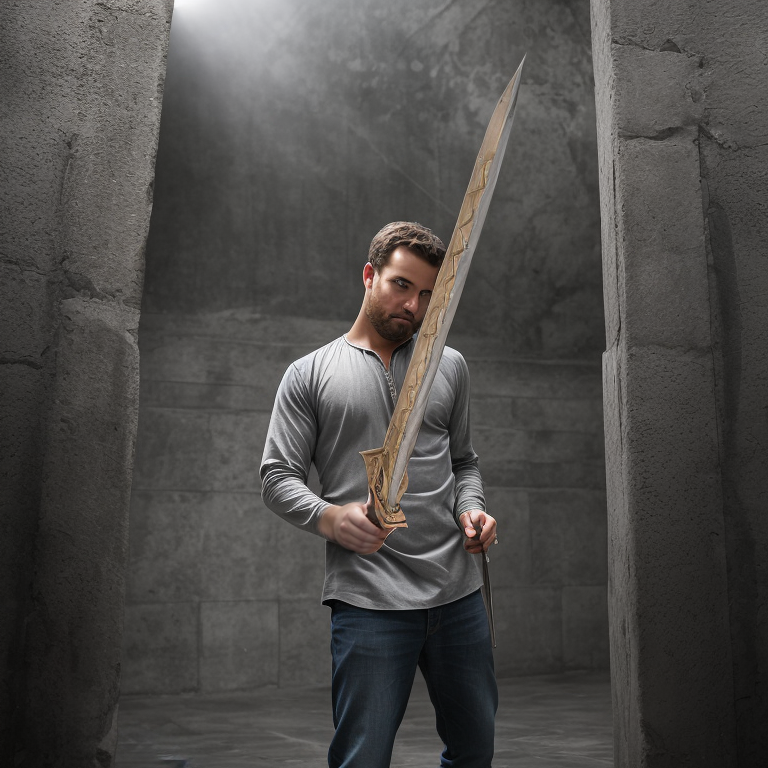}
    \end{minipage} \\
    \end{tabular}%
    }
    \vspace{-10pt}
    \caption{Failure cases for GraspDiffusion.}
    \label{fig:rebuttal2}
\end{figure}

For the second stage of our pipeline, we use a diffusion model based on the Latent Diffusion \cite{Rombach2021HighResolutionIS} architecture, and attach encoders \cite{Mou2023T2IAdapterLA} that receives spatial features from the synthesized body pose. Specifically, we first provide three spatial conditions from the full-body grasping pose; the human skeleton projection, joint depth map, and the occluded object with ambient lighting $([s^i, d^i, o^i])$. Then we further refine the hand-object region by providing similar spatial conditions, but centered on the hand region $([s_h^i, d_h^i, o_h^i])$. For training, we only train the conditional encoders and fix the parameters for the original Stable Diffusion model, encouraging the encoders to be used with other diffusion models finetuned from Stable Diffusion, accounting to our pipeline's style flexibility. 

During inference, we further control the interaction by rectifying the cross-attention maps for the human and object. For the segmentation masks from the body pose (human : $m^i$, object : $m^i_o$, negative object : $m^i_{no}$), we assign different levels of strength for each maps to create an input attention matrix $A \in \mathbf{R}^{N_i \times N_t}$, where $N_i$ and $N_t$ are the number of image and text tokens. We assign a higher weight for the object masks due to their regional size differences. We then edit the cross attention layers so that it computes the output as $\text{softmax}(\frac{QK^T+wA}{\sqrt{d_k}})V$, where $Q, K, V$ are the query, key and value embeddings, $d_k$ is the dimensionality of $Q$ and $K$, and $w$ is a scalar weight that controls the total strength of user input attention. This encourages the image tokens in the segmented regions to adhere more to the corresponding text tokens, ensuring that the interaction captured by the body pose is well maintained. Following \cite{balaji2022eDiff-I}, we calculate $w$ as 

$$
w = w^{'} \cdot \text{log}(1+\sigma) \cdot \text{max}(QK^T)
$$

where $w^{'}$ is a user defined scalar. Details on model parameters and inference are provided in Table~\ref{tab:model2params}. Note that for the feature weights, we assigned a relatively low rate for the joint depth map, to ensure the result image doesn't overfits to the SMPLX \cite{SMPL-X:2019} mesh's outline.

\section{Additional Results}
We display failure cases for our pipeline in Figure.~\ref{fig:rebuttal2}, where We note some failure cases, where the refined hand stands out from the image (left row), the hand shape tends to be uncanny (middle row), or where the complex object texture is not correctly preserved within the image (right row).

We also provide additional results for realistic, full-bodied human object interaction image generation in Figure.~\ref{fig:ab2}. We display that our model is capable of producing images of realistic humans interacting with the given object, with high diversity over human identity, body pose, camera angle, background, and other relevant scene context. The results demonstrate our pipeline's capability in creating realistic grasps for unseen objects.